\documentclass{article}
\usepackage{subcaption}
\usepackage{microtype}
\usepackage{graphicx}
\usepackage{booktabs} %
\usepackage{ragged2e}
\usepackage{float}
\usepackage[utf8]{inputenc}

\usepackage{hyperref}
\usepackage{listings}
\usepackage{float}  %
\usepackage{xcolor}

\usepackage[accepted]{icml2025}

\usepackage{amsmath}
\usepackage{amssymb}
\usepackage{mathtools}
\usepackage{amsthm}
\usepackage[capitalize,noabbrev]{cleveref}
\usepackage[inline]{enumitem}
\usepackage{multirow}

\theoremstyle{plain}

\theoremstyle{definition}

\theoremstyle{remark}

\usepackage[textsize=tiny]{todonotes}

\usepackage{adjustbox}

\icmltitlerunning{Towards Benchmarking Foundation Models for Tabular Data With Text}

\begin{document}

\twocolumn[
\icmltitle{Towards Benchmarking Foundation Models for Tabular Data With Text}

\icmlsetsymbol{equal}{*}

\begin{icmlauthorlist}
\icmlauthor{Martin Mráz}{equal,yyy}
\icmlauthor{Breenda Das}{equal,yyy}
\icmlauthor{Anshul Gupta}{equal,yyy}
\icmlauthor{Lennart Purucker}{yyy}
\icmlauthor{Frank Hutter}{pl,el,yyy}
\end{icmlauthorlist}

\icmlaffiliation{yyy}{Department of Computer Science, University of Freiburg, Freiburg, Germany}
\icmlaffiliation{pl}{Prior Labs, Freiburg, Germany}
\icmlaffiliation{el}{ELLIS Institute, Tubingen, Germany}

\icmlcorrespondingauthor{Martin Mraz}{mrazm@informatik.uni-freiburg.de}
\icmlcorrespondingauthor{Breenda Das}{dasb@informatik.uni-freiburg.de}
\icmlcorrespondingauthor{Anshul Gupta}{guptaa@informatik.uni-freiburg.de}
\icmlkeywords{Tabular Models, Benchmark, Tabular Text}
\vskip 0.3in
]

\printAffiliationsAndNotice{\icmlEqualContribution} %

\begin{abstract}

Foundation models for tabular data are rapidly evolving, with increasing interest in extending them to support additional modalities such as free-text features. However, existing benchmarks for tabular data rarely include textual columns, and identifying real-world tabular datasets with semantically rich text features is non-trivial. 
We propose a series of simple yet effective ablation-style strategies for incorporating text into conventional tabular pipelines. 
Moreover, we benchmark how state-of-the-art tabular foundation models can handle textual data by manually curating a collection of real-world tabular datasets with meaningful textual features.
Our study is an important step towards improving benchmarking of foundation models for tabular data with text.

\end{abstract}

\section{Introduction}

Foundation models have begun to transform tabular learning \citep{erickson2020autogluon-tabular, hollmann2025tabpfn}, echoing the trajectory of other research fields. 
A natural next step is mixed-modality tabular modeling, where structured columns may also include free-text fields such as job descriptions, clinical notes, or product summaries.
Current tabular benchmarks, however, almost never contain textual columns, cf. \citep{oml-benchmarking-suites, liu2024talenttabularanalyticslearning-talent, mcelfresh2024neuralnetsoutperformboosted-tabzilla}.
Moreover, locating real-world datasets with semantically rich text features is exceptionally difficult, with even exhaustive searches of OpenML and Kaggle only yielding a handful of usable candidates \citep{autogluon-text-tabular-benchmark}. 
Consequently, current tabular foundation models are rarely evaluated for tabular data with text.

Pipelines that can handle tabular data with text vary greatly in their implementation. 
AutoGluon's AutoMLPipelineFeatureGenerator ~\citep{erickson-arxiv20a}  converts text to sparse vectors; 
CARTE~\cite{kim2024cartepretrainingtransfertabular} applies fastText sentence embeddings \cite{fasttext-bojanowski2017enrichingwordvectorssubword}.
and the TabPFNv2 API accepts raw text but does not disclose its methodology.
These divergent choices raise a fundamental question: Which embedding strategy works best, and under what conditions? 

To answer this question, we present the first systematic study of predictive machine learning with foundation models for tabular data with text.
We study the performance of three representative embedding routes: fastText, Skrub’s TableVectorizer, and AutoGluon’s text encoder.
We show qualitatively, with a simple synthetic counter-example, that both n-gram based and off-the-shelf sentence embeddings can fail to recover highly predictive semantic patterns.
Moreover, quantitatively, we evaluate these methods on a manually curated set of real-world tabular benchmark that (i) contain genuinely informative free-text columns (ii) spans over a variety of domains and samples. 

\textbf{Our contributions are:}
\begin{enumerate*}[label=\textbf{(\Roman*)}]
    \item A qualitative study show casing the limitations of standard n-gram based and generic NLP-based embeddings for tabular tasks with text.
    \item  A manually curated set of real-world tabular datasets with semantically rich textual columns.
    \item  An empirical study of three text embedding pipelines for TabPFNv2 and XGBoost with the TabPFNv2 API and AutoGluon Tabular Predictor as baselines.
\end{enumerate*}

Our study reveals the limitations of current methods and underscores the need for new methods to handle tabular data with free text.

\begin{figure*}
\centering
\includegraphics[width=\textwidth]{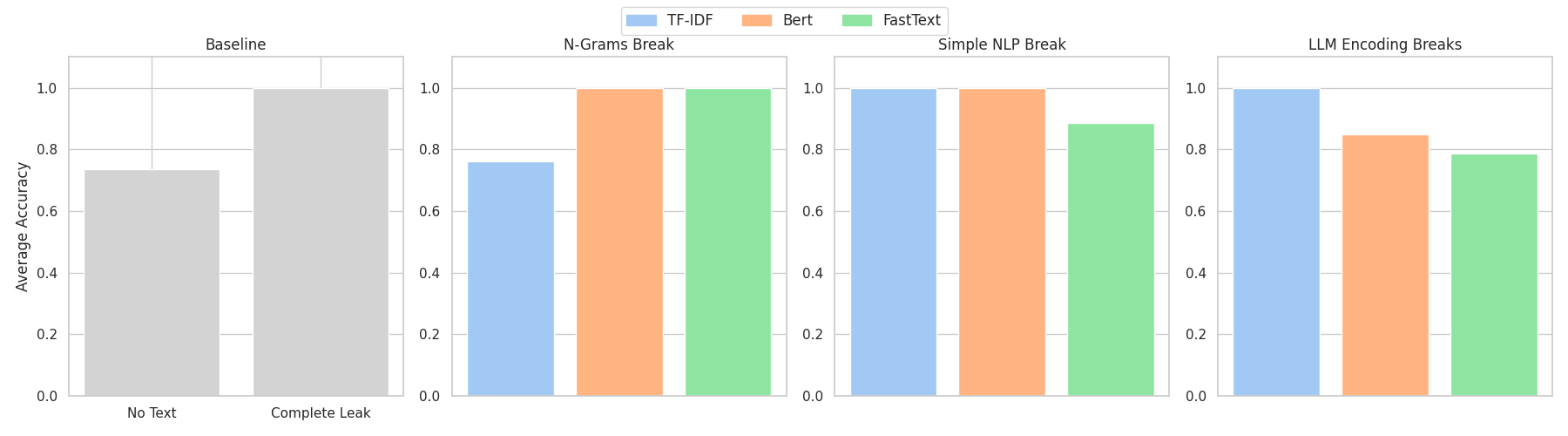}
\caption{\textbf{Qualitative investigation of textual embeddings for tabular prediction tasks:} TabPFN v2 accuracy across five datasets. Baselines ``No-Text" uses original input features, ``Complete Leak" has target leakage, therefore 100\% accuracy for all. Following tests embed targets into textual modality. ``N-Gram Break" shows TF-IDF breaking under unseen synonyms, ``Simple NLP break" shows FastText underperforming under noise, ``LLM Breaks" shows BERT variant breaking under semantic ambiguity.}

\label{motivation-plot}
\end{figure*}

\section{Related Work}
\label{sec:related}

Approaches incorporating free-text in tabular learning largely follow two paradigms.
First, \textbf{row-as-text} methods serialize entire rows into prompts and delegate prediction to a large language model (LLM), as seen in \textsc{TabLLM}~\cite{tabllm}, \textsc{Table-LLM}~\cite{tablellm}. These work well when textual fields dominate or in few-shot settings.
Second, \textbf{per-column embedding} strategies extract textual embeddings from a single or groups of features, while preserving the structural nature of tabular data.
They embed each column using fastText or LLMs and then concatenate the resulting embeddings to the table, or replace the feature with its individual textual embedding, cf. \cite {koloski2025llmembeddingsdeeplearning,carballo2023tabtextflexiblecontextualapproach, kim2024cartepretrainingtransfertabular}. 
\citet{grinsztajn2023vectorizingstringentriesdata} compare 30+ LLM and substring-based methods for embedding string columns, showing when LLMs yield better representations.

In this study, we investigate pre-column embeddings because we focus on a many-shot setting (e.g., more than 32 training samples), in which LLM-based predictive methods do not perform well, cf. \cite{hegselmann2023machinelearninghealthsymposium} and \cite{ma2024tabdptscalingtabularfoundation}, or require a prohibitive amount of fine-tuning \citep{shysheya2025joltjointprobabilisticpredictions}.

Most popular prior tabular benchmarks contain no free-text field, cf. \cite{oml-benchmarking-suites, automl-benchmark,mcelfresh-neurips23a}.
For tabular data with text, \citet{autogluon-text-tabular-benchmark} curated 18 datasets with mixed modalities, but many rely almost entirely on text, with only one or two tabular features, or features derived from the long text, e.g. \textit{\# of words} \citep{tang2024autogluonmultimodalautommsuperchargingmultimodal}.
Thus, they benchmark text-based models rather than tabular models, focusing on text data with tabular features. 
In contrast, we focus on tabular models extended to handle text, focusing on tabular data with additional text features.
Other studies were often limited to an evaluation with just one or two datasets \cite{carballo2023tabtextflexiblecontextualapproach, mug-Lu_2023}.
Overall, existing studies for tabular data with text lack a domain-agnostic benchmark where textual features complement, rather than dominate, tabular data.
This motivates the new benchmark we present.

\textbf{The CARTE Benchmark.}
The latest seminal work on benchmarking for tabular data with text is the \textsc{CARTE}~\cite{kim2024cartepretrainingtransfertabular} benchmark. 
It includes 51 datasets for tabular data with text.
However, when we investigated these datasets more closely, we found that \textbf{at most $11$ out of the $51$ datasets are suited to evaluate tabular data with text}. 
Moreover, our manually curated collection of datasets for benchmarking only includes $1$ out of the $51$ datasets. 
\\
We share an extensive report of our investigation in Appendix \ref{appendix-other-benchmnark-datasets}.
In short, we found that most datasets 
\begin{enumerate*}[label=\textbf{(\alph*)}]
    \item do not represent predictive machine learning tasks for tabular data with text;  
    \item are skewed towards text data representing many categories instead of longer free text; 
    \item were preprocessed manually per dataset with logic that seems to favor CARTE;
    \item or were very similar datasets from the same domain, with the same features, or similar semantic content. 
\end{enumerate*}
Thus, while CARTE was a significant step toward benchmarking and learning for tabular data with text, it falls short in several aspects.
Our work complements CARTE's efforts and aims to improve benchmarking for tabular data with text further.

\section{Qualitative Investigation}
\label{sec:qual}

In this section, we evaluate popular text embedding techniques: N-Grams, simple NLP models and LLMs for tabular prediction tasks. We hypothesise that N-Grams like TF-IDF break under synonym variations, simple NLP models like FastText sentence vectors \citep{fasttext-bojanowski2017enrichingwordvectorssubword} underperform under noise and LLMs like the BERT variation \cite{wang2020minilmdeepselfattentiondistillation-miniBert} might get confused under semantic ambiguity. Consider the training sentence: “The movie was bad.” and the test: “The popcorn was bad, but the movie was great.” In TF-IDF, OOD words like “good” weaken inference. FastText averages all word vectors, so noise like “popcorn was bad” disrupts the target signal. LLMs may underperform due to conflicting sentiments across long sentences. Motivated by these limitations, we design experiments to mimic such text patterns and test embedding robustness.

\textbf{Experimental Setup.} 
We choose five OpenML binary classification datasets (Appendix \ref{appendix-motivation-setup}) and formulate two baselines (see Figure \ref{motivation-plot}). For baseline “No Text” the model predicts using original input features. In baseline “Complete Leak” we leak the target feature as an addtional input feature, which results in 100\% accuracy. We test robustness of each embedding as follows: (1) We substitute the leaked binary labels with 2 semantically irrelevant words, “good” and “number” and for each sample we substitute with a synonym “great”, “three”, etc. The test split has different set of synonyms simulating OOD synonyms. (2) We substitute the binary labels with “positive” and “negative”, and populate each sample with random words e.g. - “apple mountain positive girl” simulating random noise in long text. (3) We substitute with “positive” and “negative” as before but populate each sample with opposite semantic signals e.g. - “favourable positive sad charming” simulating semantic ambiguity. Appendix~\ref{appendix-motivation-setup} has complete list of subtitutions. 

\textbf{Results and Takeaways.}
We observe in Figure \ref{motivation-plot}``N-Grams Break" subplot that TF-IDF underperforms given OOD synonyms while FastText and BERT retain 100\% accuracy demonstrating robustness to semantic variability. FastText breaks optimal performance under random noise Figure \ref{motivation-plot} ``Simple NLP Break", whereas TF-IDF and BERT retain performance, demonstrating robustness to noise under clear target signal. In Figure \ref{motivation-plot} ``LLM Breaks", semantic ambiguity obstructs BERT and FastText from identifying the target signal, whereas TF-IDF remains robust, relying on target signal frequency.

While we acknowledge that these tests may seem artificial, we also emphasise that semantic variability, random noise and semantic ambiguity are  common in real world long texts and given these limitations of prevalent embedding techniques, we might fall short of optimal performance on tabular predictions. Hence we encourage research into resolving these limitations and thereby coming closer to an ideal text embedding technique which would enable optimal tabular predictions using textual fields.

\section{Towards Quantitative Benchmarks}
\begin{table}[!h]
    \centering
    \caption{
    Overview of benchmark datasets with counts of categorical, numerical, and text columns, grouped by task type.
    }
    \vskip 0.05in
    \label{tab:column_type_stats}
    \renewcommand{\arraystretch}{1}
    \setlength\tabcolsep{3.7pt}
    \normalsize
    \centering

    \begin{tabular}{llrrrr}
    \toprule
    Dataset & Task & \# Cat & \# Num & \# Text & \# Rows \\
    \midrule
\href{https://www.kaggle.com/datasets/shivamb/real-or-fake-fake-jobposting-prediction}{fraud} & b-clf & 4 & 4 & 7 & 17880 \\
\href{https://www.kaggle.com/datasets/yashkantharia/kickstarter-campaigns}{kick} & b-clf & 10 & 4 & 4 & 94125 \\
\href{https://www.kaggle.com/datasets/ruqaiyaship/osha-accident-and-injury-data-1517}{osha} & b-clf & 15 & 1 & 3 & 4847 \\
\midrule
\href{https://www.kaggle.com/datasets/jeradrose/hearthstone-cards}{cards} & m-clf & 4 & 3 & 3 & 2810 \\
\href{https://www.kaggle.com/datasets/selener/consumer-complaint-database}{complaints} & m-clf & 4 & 2 & 5 & 96935 \\
\href{https://www.kaggle.com/datasets/maharshipandya/-spotify-tracks-dataset}{spotify} & m-clf & 5 & 11 & 3 & 10000 \\
\midrule\midrule
\href{https://www.kaggle.com/datasets/airbnb/seattle?select=listings.csv}{airbnb} & reg & 32 & 13 & 11 & 3818 \\
\href{https://www.kaggle.com/datasets/ruthgn/beer-profile-and-ratings-data-set}{beer} & reg & 3 & 13 & 5 & 2914 \\
\href{https://www.kaggle.com/competitions/california-house-prices/data?select=train.csv}{houses} & reg & 21 & 5 & 4 & 44913 \\
\href{https://www.kaggle.com/datasets/dhanushbommavaram/laptop-dataset}{laptops} & reg & 29 & 1 & 26 & 984 \\
\href{https://www.kaggle.com/c/mercari-price-suggestion-challenge/data?select=train.tsv.7z}{mercari} & reg & 2 & 1 & 4 & 100000 \\
\href{https://www.kaggle.com/datasets/aparnashastry/building-permit-applications-data}{permits} & reg & 12 & 12 & 5 & 90876 \\
\href{https://www.kaggle.com/datasets/elvinrustam/wine-dataset}{wine} & reg & 7 & 3 & 6 & 1281 \\
    \bottomrule
    \end{tabular}

\end{table}

Existing collections of datasets for benchmarking tabular data with text do not yet suffice for a robust evaluation.  
Therefore, we curate a new corpus according to five rules:
\begin{enumerate*}[label=(\roman*)]
\item \textbf{Real (free) text features.}
We select datasets that are not merely short categorical codes. Instead, we include datasets with real text features that can contain any free text.
\item \textbf{Dual-signal requirement.}  
Both textual and structural features must carry predictive information; otherwise, a task collapses into either a pure-NLP or pure-tabular test.  
\item \textbf{Tabular predictive task.}  
The dataset must be \emph{primarily} a tabular predictive task. Thus, we exclude datasets that are, for example, recommender systems or text retrieval/look-up tasks.
\item \textbf{Accessibility.}  Restricted data, such as real-world patient records (e.g.\ MIMIC-III~\cite{goldberger2000physiobank-mimicdataset}), are omitted so that the benchmark remains easily accessible for all users.
\item \textbf{Domain and target diversity.}  We do not reuse the same source datasets for multiple benchmark tasks. Instead, we cover commerce, reviews, finance, and sensor-augmented data, spanning regression and (binary/multi-class) classification tasks.  
\end{enumerate*}

Table \ref{tab:column_type_stats} overviews our curated datasets. 
Besides bootstrapping on datasets from prior work \citep{kim2024cartepretrainingtransfertabular,autogluon-text-tabular-benchmark}, we manually searched for datasets on UCI, OpenML, and Kaggle portals.
After deduplication, only Kaggle provided datasets that satisfied all four criteria. 
Pre-processing is deliberately minimal, emulating a first-pass data-scientist workflow (e.g., HTML stripping, basic imputation) and supplemented by explicit checks that no target information leaks into the inputs.  
We provide complete dataset schemas, acquisition scripts, and preprocessing details in Appendix~\ref{appendix-datasets}.
\\
In comparison to prior work, our datasets (i) exhibit a healthier balance between textual and numeric/categorical information, (ii) span diverse domains and scales,and (iii) include transparent preprocessing to ensure reproducibility.

\noindent\textit{Code availability.} 
The benchmark datasets (and more) acquisition scripts are available at \href{https://github.com/mrazmartin/TextTabBench/tree/main}{TextTabBench} repository.

\noindent\textit{A note to dataset creators.}  
We encourage the community to release tabular data \emph{before} heavy aggregation or category collapsing, so that the original free-text variation remains available for multimodal research.

\section{Quantitative Experiments and Results}
\begin{table*}[!h]
\centering
\caption{\textbf{Comparison between with-text and without-text settings:} The Table compares best accuracy obtained by text embedding techniques used in tabular models on downsampling by SHAP for each dataset against accuracy obtained with non-textual features only. For each model or framework, higher score between with-Text and without-Text is bolded.} 
\vskip 0.05in
\renewcommand{\arraystretch}{1.4}
\resizebox{\textwidth}{!}{
\begin{tabular}{l|ccc|ccc||cc cc}
\toprule
\textbf{Dataset} 
& \multicolumn{3}{c|}{\textbf{TabPFNv2}} 
& \multicolumn{3}{c||}{\textbf{XGBoost}} 
& \multicolumn{2}{c}{\textbf{TabPFNv2 API}} 
& \multicolumn{2}{c}{\textbf{AG Tabular Predictor}} 
\\
 & \textbf{Emb} & \textbf{with Text} & \textbf{w/o Text} 
 & \textbf{Emb} & \textbf{with Text} & \textbf{w/o Text} 
 & with Text & w/o Text 
 & with Text & w/o Text 
 \\
\midrule

airbnb & AGPIPE & \textbf{0.692\textsubscript{\scriptsize±0.048}} & 0.679\textsubscript{\scriptsize±0.045} & AGPIPE & \textbf{0.603\textsubscript{\scriptsize±0.060}} & 0.592\textsubscript{\scriptsize±0.048} & \textbf{0.686\textsubscript{\scriptsize±0.037}} & 0.673\textsubscript{\scriptsize±0.033} 
& 0.638\textsubscript{\scriptsize±0.023} 
& \textbf{0.645\textsubscript{\scriptsize±0.039}}
\\

beer & Fasttext & \textbf{0.646\textsubscript{\scriptsize±0.023}} & 0.579\textsubscript{\scriptsize±0.020} & Fasttext & \textbf{0.594\textsubscript{\scriptsize±0.036}} & 0.468\textsubscript{\scriptsize±0.020} & \textbf{0.643\textsubscript{\scriptsize±0.032}} & 0.573\textsubscript{\scriptsize±0.034} 
& \textbf{0.586\textsubscript{\scriptsize±0.027}} & 0.512\textsubscript{\scriptsize±0.039} 
\\

houses & Fasttext & \textbf{0.753\textsubscript{\scriptsize±0.035}} & 0.733\textsubscript{\scriptsize±0.027} & AGPIPE & \textbf{0.696\textsubscript{\scriptsize±0.020}} & 0.673\textsubscript{\scriptsize±0.028} & \textbf{0.745\textsubscript{\scriptsize±0.052}} & 0.727\textsubscript{\scriptsize±0.082} 
& 0.537\textsubscript{\scriptsize±0.154} & \textbf{0.626\textsubscript{\scriptsize±0.121}} 
\\

laptops & AGPIPE & \textbf{0.902\textsubscript{\scriptsize±0.023}} & 0.881\textsubscript{\scriptsize±0.021} & Skrub & \textbf{0.833\textsubscript{\scriptsize±0.046}} & 0.801\textsubscript{\scriptsize±0.047} & \textbf{0.900\textsubscript{\scriptsize±0.014}} & 0.868\textsubscript{\scriptsize±0.014} 
& \textbf{0.841\textsubscript{\scriptsize±0.020}} & 0.827\textsubscript{\scriptsize±0.032} 
\\

mercari & Fasttext & \textbf{0.237\textsubscript{\scriptsize±0.050}} & 0.001\textsubscript{\scriptsize±0.016} & Fasttext & \textbf{0.110\textsubscript{\scriptsize±0.062}} & 0.001\textsubscript{\scriptsize±0.006} & \textbf{0.262\textsubscript{\scriptsize±0.080}} & 0.012\textsubscript{\scriptsize±0.027} 
& \textbf{0.134\textsubscript{\scriptsize±0.037}} & -0.086\textsubscript{\scriptsize±0.095} 
\\

permits & Fasttext & 0.494\textsubscript{\scriptsize±0.062} & \textbf{0.506\textsubscript{\scriptsize±0.057}} & Fasttext & 0.426\textsubscript{\scriptsize±0.040} & \textbf{0.467\textsubscript{\scriptsize±0.065}} & \textbf{0.492\textsubscript{\scriptsize±0.050}} & 
0.470\textsubscript{\scriptsize±0.055} 
& \textbf{0.440\textsubscript{\scriptsize±0.093}}  & 0.427\textsubscript{\scriptsize±0.073}
\\

wine & Fasttext & \textbf{0.571\textsubscript{\scriptsize±0.115}} & 0.423\textsubscript{\scriptsize±0.137} & Fasttext & \textbf{0.394\textsubscript{\scriptsize±0.098}} & 0.125\textsubscript{\scriptsize±0.266} & \textbf{0.588\textsubscript{\scriptsize±0.058}} & 
0.455\textsubscript{\scriptsize±0.033} 
& \textbf{0.460\textsubscript{\scriptsize±0.064}} & 0.448\textsubscript{\scriptsize±0.054} 
\\

complaints & FastText & \textbf{0.688\textsubscript{\scriptsize±0.024}} & 0.646\textsubscript{\scriptsize±0.017} & Fasttext & \textbf{0.672\textsubscript{\scriptsize±0.021}} & 0.584\textsubscript{\scriptsize±0.023} & \textbf{0.667\textsubscript{\scriptsize±0.008}} & 0.639\textsubscript{\scriptsize±0.015} 
& \textbf{0.679\textsubscript{\scriptsize±0.011}} & 0.630\textsubscript{\scriptsize±0.024} 
\\

frauds & AGPIPE & \textbf{0.962\textsubscript{\scriptsize±0.008}} & 0.852\textsubscript{\scriptsize±0.006} & AGPIPE & \textbf{0.958\textsubscript{\scriptsize±0.004}} & 0.849\textsubscript{\scriptsize±0.015} & \textbf{0.954\textsubscript{\scriptsize±0.003}} & 
0.845\textsubscript{\scriptsize±0.012} 
& \textbf{0.960\textsubscript{\scriptsize±0.009}} & 0.848\textsubscript{\scriptsize±0.014} 
\\

cards & Fasttext & \textbf{0.703\textsubscript{\scriptsize±0.008}} & 0.662\textsubscript{\scriptsize±0.007} & Fasttext & \textbf{0.724\textsubscript{\scriptsize±0.011}} & 0.632\textsubscript{\scriptsize±0.009} & \textbf{0.705\textsubscript{\scriptsize±0.012}} & 
0.659\textsubscript{\scriptsize±0.020} 
& \textbf{0.700\textsubscript{\scriptsize±0.023}} & 0.655\textsubscript{\scriptsize±0.027} 
\\

kick & Fasttext & \textbf{0.779\textsubscript{\scriptsize±0.016}} & 0.702\textsubscript{\scriptsize±0.010} & Fasttext & \textbf{0.769\textsubscript{\scriptsize±0.014}} & 0.657\textsubscript{\scriptsize±0.013} & \textbf{0.781\textsubscript{\scriptsize±0.013}} & 0.686\textsubscript{\scriptsize±0.021} 
& \textbf{0.776\textsubscript{\scriptsize±0.020}} & 0.699\textsubscript{\scriptsize±0.020} 
\\

osha & AGPIPE & \textbf{0.613\textsubscript{\scriptsize±0.006}} & 0.566\textsubscript{\scriptsize±0.013} & AGPIPE & \textbf{0.599\textsubscript{\scriptsize±0.020}} & 0.539\textsubscript{\scriptsize±0.012} & \textbf{0.547\textsubscript{\scriptsize±0.029}} & 0.527\textsubscript{\scriptsize±0.015}
& \textbf{0.559\textsubscript{\scriptsize±0.019}} & 0.544\textsubscript{\scriptsize±0.018} \\

spotify & Fasttext & \textbf{0.815\textsubscript{\scriptsize±0.010}} & 0.663\textsubscript{\scriptsize±0.016} & Fasttext & \textbf{0.807\textsubscript{\scriptsize±0.012}} & 0.636\textsubscript{\scriptsize±0.027} & \textbf{0.841\textsubscript{\scriptsize±0.027}} & 0.665\textsubscript{\scriptsize±0.016}
& \textbf{0.735\textsubscript{\scriptsize±0.006}} & 0.636\textsubscript{\scriptsize±0.013} \\
\bottomrule
\end{tabular}
}
\label{tab:main_table}
\end{table*}

In this section we test SOTA tabular models on our dataset pool utilizing different text embedding techniques. We evaluate performance improvement obtained by each model when including textual features against dropping them.

\textbf{Experimental Setup.}
We evaluate two models, TabPFNv2~\citep{hollmann2025tabpfn} and XGBoost~\citep{chen-kdd16a}, with three text embedding methods: (1) FastText \citep{fasttext-bojanowski2017enrichingwordvectorssubword} sentence vectors, (2) Skrub’s TableVectorizer that uses GapEncoder by default, and (3) AutoGluon’s AutoMLPipelineFeatureGenerator \citep{erickson2020autogluon-tabular} that uses TextNgramFeatureGenerator. We extend these ablations to BERT embeddings \citep{wang2020minilmdeepselfattentiondistillation-miniBert} in Appendix~\ref{appendix-downsampling}. We replace the textual features in the datasets with the corresponding embeddings. Due to memory constraints on our hardware introduced by TabPFN v2 (more details in Appendix~\ref{appendix-hardware}), we evaluate various feature selection and dimensionality reduction techniques for the text embeddings, including statistical tests (t-test, ANOVA), regularization (Lasso), variance-based selection, Principal Component Analysis (PCA), correlation filtering, and SHAP-based \citep{lundberg2017unified} importance. Further downsampling details are in Appendix \ref{appendix-ds_strats}. We restrict number of features to 300 and  rows to 3000 for all evaluations, ensuring consistency. We also test TabPFNv2-API~\citep{hollmann2025tabpfn} and AutoGluon's Tabular Predictor~\citep{erickson2020autogluon-tabular} frameworks that support automatic text handling. 
We report R\textsuperscript{2} score for regression and accuracy score for classification, averaged over 5-fold cross-validation of each dataset. We do not conduct hyperparameter tuning of the text embedding generation processes for any datasets and stick to default settings. We expect performance differences upon doing so.

\textbf{Results.} In Table~\ref{tab:main_table} we compare performance of each model and framework when including textual features against not including them. In this table we report best performance amongst all embedding techniques downsampled by SHAP. 
Appendix~\ref{appendix-downsampling} shows all embedding and downsampling results.

\begin{enumerate*}[label=(\roman*)]
\item \textbf{Textual features boost accuracy.}
In 11/13 datasets, text embeddings outperforms no-text, for all models and frameworks. 

\item \textbf{No single best embedding method.}
FastText most frequently gives the best performance, on 7/13 datasets for all models and frameworks. No embedding method dominates universally.

\item \textbf{No single best dowsampling technique.}
SHAP-based selection most frequently gives the best performance, but it doesn't always dominate (detailed comparison in Appendix~\ref{appendix-downsampling}).

\item \textbf{Local models show higher gains than APIs, but APIs show consistent improvements with text.}
TabPFNv2 API consistently outperforms no-text, but local models using our default embedding pipelines often can reach higher accuracy.Hence, we can expect improvements in terms of both magnitude and consistency, and conclude text in tabular models to be an unsolved problem.
\end{enumerate*}

\textbf{Conclusion.} In conclusion we present rules for creation an effective text for tabular data benchmark and curate a pool based on these rules. We evaluate different embedding and downsampling techniques, and find that text embeddings boost performance over no-text and downsampling is crucial for effective noise mitigation and reducing memory consumption. We discuss their limitations and thereby motivate creation of a robust text-modality pipeline.

\newpage

\section*{Acknowledgements}
L.P. acknowledges funding by the Deutsche Forschungsgemeinschaft (DFG, German Research Foundation) under SFB 1597 (SmallData), grant number 499552394; and funding by ELSA – European Lighthouse on Secure and Safe AI funded by the European Union under Grant Agreement No. 101070617.
Frank Hutter acknowledges the financial support of the Hector Foundation.

\bibliography{tabadap_references,lib,proc,strings}

\begin{thebibliography}{26}
\providecommand{\natexlab}[1]{#1}
\providecommand{\url}[1]{\texttt{#1}}
\expandafter\ifx\csname urlstyle\endcsname\relax
  \providecommand{\doi}[1]{doi: #1}\else
  \providecommand{\doi}{doi: \begingroup \urlstyle{rm}\Url}\fi

\bibitem[Bischl et~al.(2019)Bischl, Casalicchio, Feurer, Hutter, Lang, Mantovani, van Rijn, and Vanschoren]{oml-benchmarking-suites}
Bischl, B., Casalicchio, G., Feurer, M., Hutter, F., Lang, M., Mantovani, R.~G., van Rijn, J.~N., and Vanschoren, J.
\newblock Openml benchmarking suites.
\newblock \emph{arXiv:1708.03731v2 [stat.ML]}, 2019.

\bibitem[Bojanowski et~al.(2017)Bojanowski, Grave, Joulin, and Mikolov]{fasttext-bojanowski2017enrichingwordvectorssubword}
Bojanowski, P., Grave, E., Joulin, A., and Mikolov, T.
\newblock Enriching word vectors with subword information, 2017.
\newblock URL \url{https://arxiv.org/abs/1607.04606}.

\bibitem[Carballo et~al.(2023)Carballo, Na, Ma, Boussioux, Zeng, Soenksen, and Bertsimas]{carballo2023tabtextflexiblecontextualapproach}
Carballo, K.~V., Na, L., Ma, Y., Boussioux, L., Zeng, C., Soenksen, L.~R., and Bertsimas, D.
\newblock Tabtext: A flexible and contextual approach to tabular data representation, 2023.
\newblock URL \url{https://arxiv.org/abs/2206.10381}.

\bibitem[Chen \& Guestrin(2016)Chen and Guestrin]{chen-kdd16a}
Chen, T. and Guestrin, C.
\newblock {XGBoost}: {A} scalable tree boosting system.
\newblock In Krishnapuram, B., Shah, M., Smola, A., Aggarwal, C., Shen, D., and Rastogi, R. (eds.), \emph{Proceedings of the 22nd {ACM} {SIGKDD} International Conference on Knowledge Discovery and Data Mining ({KDD}'16)}, pp.\  785--794, 2016.

\bibitem[Erickson et~al.(2020{\natexlab{a}})Erickson, Mueller, Shirkov, Zhang, Larroy, Li, and Smola]{erickson-arxiv20a}
Erickson, N., Mueller, J., Shirkov, A., Zhang, H., Larroy, P., Li, M., and Smola, A.
\newblock Autogluon-tabular: Robust and accurate automl for structured data.
\newblock \emph{arXiv:2003.06505 [stat.ML]}, 2020{\natexlab{a}}.

\bibitem[Erickson et~al.(2020{\natexlab{b}})Erickson, Mueller, Shirkov, Zhang, Larroy, Li, and Smola]{erickson2020autogluon-tabular}
Erickson, N., Mueller, J., Shirkov, A., Zhang, H., Larroy, P., Li, M., and Smola, A.
\newblock Autogluon-tabular: Robust and accurate automl for structured data, 2020{\natexlab{b}}.
\newblock URL \url{https://arxiv.org/abs/2003.06505}.

\bibitem[Gijsbers et~al.(2022)Gijsbers, Bueno, Coors, LeDell, Poirier, Thomas, Bischl, and Vanschoren]{automl-benchmark}
Gijsbers, P., Bueno, M. L.~P., Coors, S., LeDell, E., Poirier, S., Thomas, J., Bischl, B., and Vanschoren, J.
\newblock Amlb: an automl benchmark, 2022.
\newblock URL \url{https://arxiv.org/abs/2207.12560}.

\bibitem[Goldberger et~al.(2000)Goldberger, Amaral, Glass, Hausdorff, Ivanov, Mark, Mietus, Moody, Peng, and Stanley]{goldberger2000physiobank-mimicdataset}
Goldberger, A.~L., Amaral, L.~A., Glass, L., Hausdorff, J.~M., Ivanov, P.~C., Mark, R.~G., Mietus, J.~E., Moody, G.~B., Peng, C.-K., and Stanley, H.~E.
\newblock Physiobank, physiotoolkit, and physionet: Components of a new research resource for complex physiologic signals.
\newblock \emph{Circulation}, 101\penalty0 (23):\penalty0 e215--e220, 2000.

\bibitem[Grinsztajn et~al.(2023)Grinsztajn, Oyallon, Kim, and Varoquaux]{grinsztajn2023vectorizingstringentriesdata}
Grinsztajn, L., Oyallon, E., Kim, M.~J., and Varoquaux, G.
\newblock Vectorizing string entries for data processing on tables: when are larger language models better?, 2023.
\newblock URL \url{https://arxiv.org/abs/2312.09634}.

\bibitem[Hegselmann et~al.(2022)Hegselmann, Buendia, Lang, Agrawal, Jiang, and Sontag]{tabllm}
Hegselmann, S., Buendia, A., Lang, H., Agrawal, M., Jiang, X., and Sontag, D.
\newblock Tabllm: Few-shot classification of tabular data with large language models, 2022.
\newblock URL \url{https://arxiv.org/abs/2210.10723}.

\bibitem[Hegselmann et~al.(2023)Hegselmann, Parziale, Shanmugam, Tang, Asiedu, Chang, Hartvigsen, and Singh]{hegselmann2023machinelearninghealthsymposium}
Hegselmann, S., Parziale, A., Shanmugam, D., Tang, S., Asiedu, M.~N., Chang, S., Hartvigsen, T., and Singh, H.
\newblock Machine learning for health symposium 2023 -- findings track, 2023.
\newblock URL \url{https://arxiv.org/abs/2312.00655}.

\bibitem[Hollmann et~al.(2025)Hollmann, M{\"u}ller, Purucker, Krishnakumar, K{\"o}rfer, Hoo, Schirrmeister, and Hutter]{hollmann2025tabpfn}
Hollmann, N., M{\"u}ller, S., Purucker, L., Krishnakumar, A., K{\"o}rfer, M., Hoo, S.~B., Schirrmeister, R.~T., and Hutter, F.
\newblock Accurate predictions on small data with a tabular foundation model.
\newblock \emph{Nature}, 01 2025.
\newblock \doi{10.1038/s41586-024-08328-6}.
\newblock URL \url{https://www.nature.com/articles/s41586-024-08328-6}.

\bibitem[Kim et~al.(2024)Kim, Grinsztajn, and Varoquaux]{kim2024cartepretrainingtransfertabular}
Kim, M.~J., Grinsztajn, L., and Varoquaux, G.
\newblock Carte: Pretraining and transfer for tabular learning, 2024.
\newblock URL \url{https://arxiv.org/abs/2402.16785}.

\bibitem[Koloski et~al.(2025)Koloski, Margeloiu, Jiang, Škrlj, Simidjievski, and Jamnik]{koloski2025llmembeddingsdeeplearning}
Koloski, B., Margeloiu, A., Jiang, X., Škrlj, B., Simidjievski, N., and Jamnik, M.
\newblock Llm embeddings for deep learning on tabular data, 2025.
\newblock URL \url{https://arxiv.org/abs/2502.11596}.

\bibitem[Liu et~al.(2024)Liu, Cai, Zhou, and Ye]{liu2024talenttabularanalyticslearning-talent}
Liu, S.-Y., Cai, H.-R., Zhou, Q.-L., and Ye, H.-J.
\newblock Talent: A tabular analytics and learning toolbox.
\newblock \emph{arXiv preprint arXiv:2407.04057}, 2024.

\bibitem[Lu et~al.(2023)Lu, Qian, Zhao, Xi, and Yang]{mug-Lu_2023}
Lu, J., Qian, Y., Zhao, S., Xi, Y., and Yang, C.
\newblock Mug: A multimodal classification benchmark on game data with tabular, textual, and visual fields.
\newblock In \emph{Findings of the Association for Computational Linguistics: EMNLP 2023}, pp.\  5332–5346. Association for Computational Linguistics, 2023.
\newblock \doi{10.18653/v1/2023.findings-emnlp.354}.
\newblock URL \url{http://dx.doi.org/10.18653/v1/2023.findings-emnlp.354}.

\bibitem[Lundberg \& Lee(2017)Lundberg and Lee]{lundberg2017unified}
Lundberg, S.~M. and Lee, S.-I.
\newblock A unified approach to interpreting model predictions.
\newblock \emph{Advances in neural information processing systems}, 30, 2017.

\bibitem[Ma et~al.(2024)Ma, Thomas, Hosseinzadeh, Kamkari, Labach, Cresswell, Golestan, Yu, Volkovs, and Caterini]{ma2024tabdptscalingtabularfoundation}
Ma, J., Thomas, V., Hosseinzadeh, R., Kamkari, H., Labach, A., Cresswell, J.~C., Golestan, K., Yu, G., Volkovs, M., and Caterini, A.~L.
\newblock Tabdpt: Scaling tabular foundation models, 2024.
\newblock URL \url{https://arxiv.org/abs/2410.18164}.

\bibitem[Maćkiewicz \& Ratajczak(1993)Maćkiewicz and Ratajczak]{pca-article}
Maćkiewicz, A. and Ratajczak, W.
\newblock Principal components analysis (pca).
\newblock \emph{Computers \& Geosciences}, 19\penalty0 (3):\penalty0 303--342, 1993.
\newblock ISSN 0098-3004.
\newblock \doi{https://doi.org/10.1016/0098-3004(93)90090-R}.
\newblock URL \url{https://www.sciencedirect.com/science/article/pii/009830049390090R}.

\bibitem[McElfresh et~al.(2023)McElfresh, Khandagale, Valverde, {Prasad C}, Ramakrishnan, Goldblum, and White]{mcelfresh-neurips23a}
McElfresh, D., Khandagale, S., Valverde, J., {Prasad C}, V., Ramakrishnan, G., Goldblum, M., and White, C.
\newblock When do neural nets outperform boosted trees on tabular data?
\newblock In \emph{Proceedings of the 37th International Conference on Advances in Neural Information Processing Systems ({N}eur{IPS}'23)}, pp.\  76336--76369, 2023.

\bibitem[McElfresh et~al.(2024)McElfresh, Khandagale, Valverde, C, Feuer, Hegde, Ramakrishnan, Goldblum, and White]{mcelfresh2024neuralnetsoutperformboosted-tabzilla}
McElfresh, D., Khandagale, S., Valverde, J., C, V.~P., Feuer, B., Hegde, C., Ramakrishnan, G., Goldblum, M., and White, C.
\newblock When do neural nets outperform boosted trees on tabular data?, 2024.
\newblock URL \url{https://arxiv.org/abs/2305.02997}.

\bibitem[Shi et~al.(2021)Shi, Mueller, Erickson, Li, and Smola]{autogluon-text-tabular-benchmark}
Shi, X., Mueller, J., Erickson, N., Li, M., and Smola, A.~J.
\newblock Benchmarking multimodal automl for tabular data with text fields, 2021.
\newblock URL \url{https://arxiv.org/abs/2111.02705}.

\bibitem[Shysheya et~al.(2025)Shysheya, Bronskill, Requeima, Siddiqui, Gonzalez, Duvenaud, and Turner]{shysheya2025joltjointprobabilisticpredictions}
Shysheya, A., Bronskill, J., Requeima, J., Siddiqui, S.~A., Gonzalez, J., Duvenaud, D., and Turner, R.~E.
\newblock Jolt: Joint probabilistic predictions on tabular data using llms, 2025.
\newblock URL \url{https://arxiv.org/abs/2502.11877}.

\bibitem[Tang et~al.(2024)Tang, Fang, Zhou, Yang, Zhong, Hu, Kirchhoff, and Karypis]{tang2024autogluonmultimodalautommsuperchargingmultimodal}
Tang, Z., Fang, H., Zhou, S., Yang, T., Zhong, Z., Hu, T., Kirchhoff, K., and Karypis, G.
\newblock Autogluon-multimodal (automm): Supercharging multimodal automl with foundation models, 2024.
\newblock URL \url{https://arxiv.org/abs/2404.16233}.

\bibitem[Wang et~al.(2020)Wang, Wei, Dong, Bao, Yang, and Zhou]{wang2020minilmdeepselfattentiondistillation-miniBert}
Wang, W., Wei, F., Dong, L., Bao, H., Yang, N., and Zhou, M.
\newblock Minilm: Deep self-attention distillation for task-agnostic compression of pre-trained transformers, 2020.
\newblock URL \url{https://arxiv.org/abs/2002.10957}.

\bibitem[Zhang et~al.(2024)Zhang, Luo, Zhang, Ma, Zhang, Li, Li, Yao, Xu, Zhou, Zhang-Li, Yu, Zhao, Li, and Tang]{tablellm}
Zhang, X., Luo, S., Zhang, B., Ma, Z., Zhang, J., Li, Y., Li, G., Yao, Z., Xu, K., Zhou, J., Zhang-Li, D., Yu, J., Zhao, S., Li, J., and Tang, J.
\newblock Tablellm: Enabling tabular data manipulation by llms in real office usage scenarios, 2024.
\newblock URL \url{https://arxiv.org/abs/2403.19318}.

\end{thebibliography}
\bibliographystyle{icml2025}

\newpage
\appendix
\onecolumn

\section{TextTabBench Datasets}
\label{appendix-datasets}

We applied a consistent preprocessing pipeline to all datasets prior to training, aiming to ensure quality, reduce noise, and standardize formats across diverse sources.

\paragraph{General Preprocessing.}
After downloading the raw CSV files, each dataset was cropped to a maximum of \textbf{100{,}000 rows} to reduce memory and compute overhead. This limit was chosen as it extensively exceeds the typical size needed to train a performant in-context learner in our setting.

We then applied the following preprocessing steps:
\begin{itemize}
    \item \textbf{Missing value filtering:} Columns with more than 50\% missing values were dropped
    (except in a few borderline cases)
    \item \textbf{Constant column removal:} Columns containing only a single unique value (often placeholder IDs or corrupted entries) were removed.
    \item \textbf{Duplicate removal:} Exact duplicate rows were dropped to avoid sampling bias.
    \item \textbf{Target validation:} Rows with missing target labels were discarded.
    \item \textbf{Unnamed column cleanup:} Spurious columns labeled ``Unnamed'' (typically from CSV formatting issues) were dropped.
\end{itemize}

\paragraph{Column Type Classification.}
To determine appropriate encoding strategies, we automatically classified columns into \emph{numerical}, \emph{categorical}, and \emph{textual} types using the following heuristics:

\begin{itemize}
    \item \textbf{Numerical columns:}
    \begin{itemize}
        \item Columns explicitly typed as numeric were preserved.
        \item String columns were also considered numerical if their values were mostly numeric after stripping relatively few (towards total char length) formatting characters (e.g., ``15s'', $\rightarrow$ 15).
        \item Additionally, if the non-numeric part of a string column was repetitive across rows (e.g., [``ABV 12\%'', ``ABV 15\%'', ``ABV 10\%''] $\rightarrow$ [12, 15, 10]), the column was interpreted as numeric.
        \item After being classified as \textit{numerical}, these columns were automatically cleaned of non-numeric artefacts.
    \end{itemize}
    
    \item \textbf{Categorical columns:}
    \begin{itemize}
        \item The threshold for categoricity was typically set to 50 unique values for large datasets, or computed as 5\% of the number of rows for smaller ones.
        \item In rare cases, this boundary was manually adjusted, for example, when the number of unique values slightly exceeded the default threshold (e.g., 51 categories in a dataset with several thousand rows).
    \end{itemize}

    \item \textbf{Textual columns:}
    \begin{itemize}
        \item String columns with a large number of unique values and little to no numeric structure were classified as textual.
    \end{itemize}
\end{itemize}

These heuristics ensured robust and interpretable categorisation across the heterogeneous collection of datasets. The category assignments were manually verified for correctness, and in a small number of cases required reassessment and tailored adjustments. For example, string placeholders like ``no-data'' were mapped to \texttt{NaN}, zip codes embedded in longer strings (e.g., ``1234XXX'') were trimmed to their numeric prefix (``1234''). In features that were predominantly numerical, if only a few rows contained irregular textual entries (e.g. ``3 or 4, not sure''), these were manually cleaned and converted to approximate numeric values (e.g., ``3.5'') rather than dropping the entire feature, to preserve data utility.

\paragraph{Manual Dataset-Specific Adjustments.}
Following the general preprocessing and type classification, we applied an additional round of light manual cleaning. These modifications were dataset-specific and aimed at removing irrelevant or non-essential columns (e.g., constant metadata, indices, or redundant identifiers), which could otherwise introduce noise or imbalance.

To ensure fairness across model types, we explicitly avoided transformations that would favor a particular modality. For instance, we \emph{did not} split compound textual columns into multiple fields, as such engineering might disproportionately benefit text-aware models. 

The types of changes applied in this step included:
\begin{itemize}
    \item Dropping hand-identified non-informative columns (IDs, replicated features, etc.).
    \item Converting clearly timestamp-formatted strings into UNIX numeric time representations.
    \item Personalised preprocessing of numerical features (e.g. a conversion ['700ml', '8dl', '0.65l', ...] $\rightarrow$ [700, 800, 650, ...])
\end{itemize}

These dataset-specific adjustments were kept minimal and only made when necessary to ensure compatibility, reduce noise, or preserve valid numerical representations, without imposing any domain expertise.We can summarise these choices as "What would any reasonable person do, given no knowledge about which model is to be used over the data?"

\paragraph{Reproducibility.}
All such dataset-specific changes are clearly documented in the corresponding preprocessing notebooks within our benchmark repository.
The code to reproduce our results is to be found at
\href{https://github.com/mrazmartin/TowardsTextTabBench/tree/main}{Towards TextTabBench} repository.

\subsection*{Classification Datasets}

\begin{itemize}

  \item \textbf{Consumer Complaints Dataset}
  This dataset contains records of consumer complaints filed against financial service companies, featuring structured fields such as product type, state, and submission method, alongside rich textual attributes like sub-product, issue, sub-issue, and company name. The prediction target is the company's response to the complaint, consolidated into four meaningful classes: \textsc{Closed with explanation, Closed with non-monetary relief, Closed with monetary relief, and Closed without relief}. This focused setup maintains a challenging multi-class classification task while allowing for a fair assessment of how textual embeddings can enhance tabular model performance in real-world customer service scenarios. The original dataset has approx. 1.42m rows, which we randomly downsample to 100k to include a fair representation of rare classes while keeping the size feasible for training.

  \item \textbf{Hearthstone Cards Dataset}

  Describes collectable cards from the game \textit{Hearthstone}, with features covering gameplay stats, categorical traits, and rich textual descriptions. The prediction target is the \textit{player class} to which a card belongs (10 categories), with a notable class imbalance. \textit{Neutral} cards dominate, while other classes have significantly fewer samples. This natural imbalance reflects the game design and poses a realistic challenge for classification tasks. Text fields like \textit{text} (card effects) and \textit{flavor} (lore snippets) provide valuable semantic signals, making this dataset well-suited to benchmark how textual features can improve tabular model performance, especially for minority classes.

  \item \textbf{Job Posting Fraud Detection}

  This dataset contains job advertisements with structured attributes (e.g., employment type, required experience) and rich textual content (e.g., job title, description, company profile). The prediction target is \textit{fraudulent} (binary classification: real vs fake). Textual fields such as \textit{description}, \textit{requirements}, and \textit{company\_profile} carry essential semantic cues for detecting fraudulent postings, making this dataset highly suitable for evaluating text-augmented tabular models. The target is notably imbalanced, with legitimate postings outnumbering fraudulent ones.

  \item \textbf{Spotify Genres Dataset}

  This dataset contains audio features and metadata for a large collection of songs, combining numerical attributes (e.g., danceability, tempo), categorical musical traits (e.g., key, time signature), and rich textual fields such as \textit{track name}, \textit{album name}, and \textit{artist}. The prediction target is \textit{track genre}, we downsample the original 114 balanced genres to only 10, easier to differentiate classes:
  
  \textsc{['classical', 'country', 'electronic', 'hip-hop', 'indie', 'jazz', 'metal', 'pop', 'r-n-b', 'rock']}

  \item \textbf{Kickstarter Success Dataset}

  This dataset contains metadata from Kickstarter crowdfunding projects, including campaign characteristics such as \textit{funding goal}, \textit {duration}, and \textit{category information}, as well as descriptive textual fields like \textit{project name}, \textit{sub-category}, and \textit{city}. The prediction target is the binary \textit{status} (successful vs. failed). Temporal features like \textit{launched\_at} and \textit{deadline} are represented as numerical timestamps.

  \item \textbf{OSHA Accidents Dataset}

  This dataset captures detailed records of workplace accidents reported to the Occupational Safety and Health Administration (OSHA), combining structured attributes such as \textit{project type}, \textit{degree of injury}, and \textit{part of body affected} with rich textual descriptions including \textit{abstract summaries}, \textit{event descriptions}, and \textit{associated keywords}. The prediction target is whether the injured employee was performing a \textit{regularly assigned task} at the time of the incident (binary classification).

\end{itemize}

\subsection*{Regression Datasets}

\begin{itemize}
  \item \textbf{Airbnb Rates Dataset} \\
  This dataset includes Airbnb listings from Seattle with structured features such as \textit{property type}, \textit{room type}, \textit{coordinates}, and \textit{host metrics}, alongside rich textual fields like \textit{listing summaries}, \textit{neighborhood overviews}, and \textit{amenities}. The prediction target is the \textit{listing price} (continuous regression). It is a mix of structured and textual inputs making it an excellent benchmark for assessing how well tabular models can leverage textual embeddings in price prediction tasks.

  \item \textbf{California Housing Prices} \\
  The dataset includes structured property details such as \textit{lot size}, \textit{year built}, and \textit{school quality}, along with free-text descriptions of \textit{home features} (e.g., \textit{appliances}, \textit{flooring}, \textit{heating}). The prediction target is \textit{Total Interior Livable Area}, shifting focus away from price toward physical home size estimation. Price-related columns are excluded to prevent leakage. This setting tests how well textual features help infer structural characteristics.

  \item \textbf{Laptop Dataset} \\
  This dataset contains laptop listings with \textit{specifications}, \textit{textual descriptions}, and target prices. Inputs include categorical traits (e.g., \textit{Processor Brand}), numerical specs (e.g., \textit{RAM}, \textit{Screen Size}), and verbose text (e.g., \textit{Sales Package}, \textit{Additional Features}). Rather than filtering, all features are retained to simulate real-world noise and redundancy. It serves as a robust stress test for tabular foundation models under messy conditions.

  \item \textbf{Mercari Price Suggestion Dataset} \\
  This dataset comes from a consumer-to-consumer marketplace and includes listings with structured fields (e.g., \textit{item condition}, \textit{shipping}) and high-cardinality textual inputs like \textit{product names}, \textit{brand names}, and \textit{descriptions}. The goal is to predict the \textit{item price}. Hierarchical category fields (e.g., \textit{Women/Jewelry/Necklaces}) are retained as textual inputs. The dataset is ideal for evaluating semantic reasoning in text-rich, user-generated data.

  \item \textbf{San Francisco Building Permits Dataset} \\
  The task is to predict \textit{time\_to\_approve} (in days) for permit applications. Features include structured fields (e.g., \textit{permit type}, \textit{construction type}, \textit{estimated cost}) and descriptive metadata (e.g., \textit{project description}, \textit{location}). While many permits have zero delay (often legitimate), these are retained. The target’s long-tail distribution together with high frequency of zero-day-permits poses a meaningful challenge for modelling bureaucratic latency with mixed inputs.

  \item \textbf{Wine Cost Dataset} \\
  This dataset includes structured product data (e.g., \textit{grape variety}, \textit{vintage}, \textit{ABV}) and rich textual descriptions such as sensory profiles and marketing text. The prediction target is \textit{price}. Given the subjective nature of descriptions and influence on perceived value, this setting benchmarks how well tabular models can integrate free text for subjective value estimation.

\end{itemize}

\section{Downsampling Strategies}
\label{appendix-ds_strats}

We select a diverse set of feature downsampling techniques to evaluate across datasets and embedding strategies.  
These include both \textit{supervised} methods that rely on the target variable and \textit{unsupervised} methods that operate independently of it.  
Some techniques are task-specific—for example, statistical tests require classification targets, and correlation-based filters are only meaningful for regression.  
They also vary in complexity: while some are zero-shot filters based on simple statistics (e.g., variance, correlation), others require training a predictive model to estimate feature importance (e.g., SHAP, L1 regularization).  
This diversity allows us to compare the effectiveness of different selection strategies under varied conditions and modeling setups.

\newcommand{\cmark}{\checkmark}  %
\newcommand{\xmark}{\texttimes}  %

\begin{table}[h!]
\centering
\caption{
\textbf{Applicability and properties of feature selection strategies. }A checkmark (\cmark) indicates support or applicability; a cross (\xmark) indicates that the method is not suitable for that task type or does not have the specified property.
}
\vskip 0.05in
\renewcommand{\arraystretch}{1.2}
\begin{tabular}{l|ccc|cc}
\toprule
\textbf{Selector} & \textbf{Regression} & \textbf{Binary Clf} & \textbf{Multi-class Clf} & \textbf{Zero-Shot} & \textbf{Supervised} \\
\midrule
\textit{t-test}         & \xmark & \cmark & \xmark & \cmark & \cmark \\
\textit{ANOVA}          & \xmark & \cmark & \cmark & \cmark & \cmark \\
\textit{L1 (Lasso)}     & \cmark & \cmark & \cmark & \xmark & \cmark \\
\textit{Variance}       & \cmark & \cmark & \cmark & \cmark & \xmark \\
\textit{PCA}            & \cmark & \cmark & \cmark & \cmark & \xmark \\
\textit{Correlation}    & \cmark & \xmark & \xmark & \cmark & \cmark \\
\textit{SHAP}           & \cmark & \cmark & \cmark & \xmark & \cmark \\
\textit{Random}         & \cmark & \cmark & \cmark & \cmark & \xmark \\
\bottomrule
\end{tabular}

\label{tab:selector-applicability}
\end{table}

\paragraph{T-test}
This method checks whether the average value of each feature is different between binary classes.  
It runs a statistical test (t-test) on each feature and selects those with the most significant differences.  
Only works for binary classification. No model training is needed.

\paragraph{ANOVA}
This method checks if the average values of a feature are different across three or more classes.  
It runs a statistical test (ANOVA) on each feature and keeps the ones with the strongest differences.  
Works for both binary and multi-class classification. No model training is needed.

\paragraph{Variance}
This method selects features that vary the most across all rows.  
It keeps the top-\(k\) features with the highest variance, assuming that more variable features may carry more information.  
It is unsupervised and does not use the target variable.

\paragraph{PCA} \cite{pca-article}
This method uses Principal Component Analysis (PCA) to find combinations of features that explain how the data varies the most.  
Each feature gets a score based on how much it contributes to these combinations.  
This score is computed by taking the feature’s contribution (called a loading), multiplying it by how much that combination explains, and summing across all combinations.  
The features with the highest total scores are kept. This method does not use the target variable.

\paragraph{L1 regularisation}
This method trains a linear model with an \(\ell_1\) penalty, which forces some feature weights to become exactly zero.  
For regression tasks, it uses Lasso (linear regression with L1 regularisation).  
For classification, it uses logistic regression with an L1 penalty.  
After training, it selects the top-\(k\) features with the largest non-zero weights.  
This method is supervised and requires fitting a model.

\textit{Disclaimer:} We added this downsampling method only for the additional BERT embeddings experiment. 

\paragraph{Correlation}
This method computes the correlation between each feature and the target, using either Pearson or Spearman correlation.  
It ranks features by the absolute correlation and selects the top-\(k\).  
It is supervised and mainly used for regression. It can also be applied to binary classification, but is less reliable than methods like the t-test.

\paragraph{SHAP} \cite{lundberg2017unified}
This method trains a model and then uses SHAP values to measure how much each feature contributes to the model’s predictions.  
It computes the average absolute SHAP value for each feature across the dataset and selects the top-\(k\) most important ones.  
It is supervised and requires training a model such as XGBoost or logistic regression.

\section{Hyperparameters and Training Details}

\label{appendix-hparams}

\subsection*{Training Settings Summary}

\textbf{TabPFNv2:} The TabPFNv2 model is used in inference-only mode with the argument \texttt{ignore\_pretraining\_limits=True}, allowing it to bypass dataset constraints from pretraining. 

\textbf{XGBoost:} The XGBoost model uses default hyperparameters unless otherwise specified. For binary classification, it uses the \texttt{binary:logistic} objective and \texttt{logloss} as the evaluation metric. For multi-class classification, it uses \texttt{multi:softprob} with \texttt{mlogloss}. Regression uses the default objective (\texttt{reg:squarederror}). Label encoding is disabled via \texttt{use\_label\_encoder=False}. Early stopping, max depth, and learning rate are not explicitly set, so the defaults apply (\texttt{max\_depth=6}, \texttt{learning\_rate=0.3}).

\textbf{TabPFNv2 API:} This refers to the standardized inference interface provided in the TabPFN \cite{hollmann2025tabpfn}, which enables users to leverage the pretrained TabPFNv2 model without the need of external hardware like GPU. For best results, we kept similar samples to other experiments and ensure that all datasets were passed as raw tables (with basic preprocessing as already stated in the paper except for `osha' where the API does not support ``Abstract Text" column and crashes) to the API in our experiments with and without text.
    
\textbf{AutoGluon's Tabular Predictor:} It is included as a strong AutoML baseline.  It is trained per fold with \texttt{presets=best quality} and \texttt{time\_limit=360} seconds across most datasets on \texttt{eval\_metrics} of R\textsuperscript{2} score for regression and \texttt{accuracy} for classification tasks. These adjustments allowed AutoGluon's internal model selection, hyperparameter tuning, and ensembling mechanisms to perform more effectively and fairly reflect the model’s potential. 

Our time constraints were selected to ensure computational feasibility and comparability across datasets. A more generous budget could allow these systems to explore a richer search space and potentially yield higher predictive performance, however we prefer shorter budgets to compare to the ICL performance of TabPFN. Conducting a fair comparison would then necessitate tuning TabPFN as well—an approach that has already demonstrated superior performance compared to AutoGluon's Tabular Predictor under default configurations.

\section{Hardware Specifications}

\label{appendix-hardware}

We ran all experiments on a single NVIDIA GeForce
RTX 2080 Ti GPU with 11 GB RAM for each experiment on TabPFNv2  and single CPU core with 6 GB RAM for each XGBoost experiment, and 32GB RAM for AutoGluon's Tabular Predictor.

\begin{table}[!h]
\caption{Classification accuracies across break points and embedding methods.}
\label{combined-breaks-table}
\vskip 0.15in
\begin{center}
\begin{small}
\begin{sc}
\begin{tabular}{llccc}
\toprule
Break Type & Dataset ID & TF-IDF & BERT & FastText (FT) \\
\midrule
\multirow{4}{*}{TF-IDF Breaks} 
    & 31    & 80.0   & 100.0 & 100.0 \\
    & 42193 & 55.0   & 100.0 & 100.0 \\
    & 1461  & 80.0   & 100.0 & 100.0 \\
    & 1590  & 89.5   & 100.0 & 100.0 \\
\cmidrule{2-5}
    & Average & 76.1 & 100.0 & 100.0 \\
\midrule
\multirow{4}{*}{FastText Breaks} 
    & 31    & 100.0  & 100.0 & 85.0 \\
    & 42193 & 100.0  & 100.0 & 95.0 \\
    & 1461  & 100.0  & 100.0 & 80.0 \\
    & 1590  & 100.0  & 100.0 & 94.7 \\
\cmidrule{2-5}
    & Average & 100.0 & 100.0 & 88.7 \\
\midrule
\multirow{4}{*}{BERT Breaks} 
    & 31    & 100.0  & 85.0  & 75.0 \\
    & 42193 & 100.0  & 80.0  & 70.0 \\
    & 1461  & 100.0  & 80.0  & 80.0 \\
    & 1590  & 100.0  & 94.7  & 89.5 \\
\cmidrule{2-5}
    & Average & 100.0 & 84.9 & 78.6 \\
\bottomrule
\end{tabular}
\end{sc}
\end{small}
\end{center}
\vskip -0.1in
\end{table}

\section{Motivational Setup}
\label{appendix-motivation-setup}
\textbf{Datasets Used}: We used four OpenML datasets for aggregating the scores of the three tested embedding techniques for the motivational setup. We subsample 100 rows randomly in each dataset for these test scenarios. The table \ref{combined-breaks-table} details the scores for each dataset and embedding. 

\textbf{Complete list of Dilution Strategies.} We present the complete lists of target replacement strategies used in the three test scenarios (excluding the baselines). The full set of synonyms for the terms 'good' and 'number' used in the TF-IDF breakpoint simulation is shown in Listing~\ref{lst:tfidf-breakpoint}. The full set of random words used for the FastText breakpoint simulation appears in Listing~\ref{lst:ft-breakpoint}. The complete list of synonyms appended to 'positive' or 'negative' in the Bert-based test is provided in Listing~\ref{lst:bert-breakpoint}.

\definecolor{codebg}{HTML}{F7F7F7}
\definecolor{commentgray}{HTML}{999988}
\definecolor{keywordblue}{HTML}{0000AA}
\definecolor{stringred}{HTML}{A31515}
\definecolor{identifierblack}{HTML}{333333}

\lstdefinestyle{mystyle}{
  backgroundcolor=\color{codebg},
  basicstyle=\ttfamily\small,
  keywordstyle=\color{keywordblue}\bfseries,
  stringstyle=\color{stringred},
  commentstyle=\color{commentgray}\itshape,
  identifierstyle=\color{identifierblack},
  showstringspaces=false,
  numberstyle=\tiny\color{gray},
  numbers=left,
  stepnumber=1,
  numbersep=5pt,
  frame=single,
  breaklines=true,
  breakatwhitespace=true,
  captionpos=b
}
\lstset{style=mystyle}

\begin{minipage}{0.95\linewidth}
\begin{lstlisting}[language=Python, numbers=none, caption={TF-IDF Breakpoint Simulation}, 
label={lst:tfidf-breakpoint}]
self.synonyms = {
    'train': {
        "good": [
            "positive", "great", "excellent", "favorable", "pleasant",
            "admirable", "beneficial", "wonderful", "commendable", "worthy"
        ],
        "number": [
            "one", "three", "four", "five", "six",
            "seven", "eight", "nine", "ten", "eleven"
        ]
    },
    'test': {
        "good": ["nice"],
        "number": ["two"]
    }
}
\end{lstlisting}
\end{minipage}

\begin{minipage}{0.95\linewidth}
\begin{lstlisting}[language=Python, numbers=none, caption={FastText Breakpoint Simulation}, label={lst:ft-breakpoint}]
 self.random_words = [
    "breeze", "crystal", "jungle", "sunset", "clock",
    "river", "pencil", "butterfly", "cloud", "guitar", "forest",
    "echo", "mirror", "flame", "galaxy", "shadow", "storm", "pearl",
    "ember", "whisper", "velvet", "feather", "lantern", "cherry", "fog",
    "nutmeg", "rocket", "canyon", "harbor", "planet", "sketch", "compass",
    "dream", "saddle", "maple", "python", "quartz", "cactus", "ladder", 
    "amber", "panther", "blanket", "marble", "candle", "helmet",
    "anchor", "sand", "ocean", "lemon", "boulder", "ink", "ribbon",
    "nest", "basket", "flute", "meadow", "thunder", "vine", "shell",
    "drift", "carpet", "sapphire", "tiger", "honey", "blossom", "stream",
    "mountain", "lighthouse", "cliff", "pebble", "tunnel", "bubble",
    "apple", "silver", "chalk", "frost", "comet", "antler", "bramble",
    "ripple", "beacon", "groove", "hazel", "dune", "harvest",
    "twig", "cobweb", "glider", "ivory", "petal",  "plume",
    "island", "whistle", "puzzle", "snowflake", "cradle",
    "nail", "window", "tassel"
 ]

\end{lstlisting}
\end{minipage}

\begin{minipage}{0.95\linewidth}
\begin{lstlisting}[language=Python, numbers=none, caption={Bert Breakpoint Simulation},
label={lst:bert-breakpoint}]
 self.random_words =  [           
    # Positive Synonyms
    "favorable", "happy", "joyful", "pleased", "delighted", "cheerful", "content", "grateful", "optimistic", "upbeat", "ecstatic",
    "radiant", "thrilled", "hopeful", "enthusiastic", "elated", "blissful", "satisfied", "charming", "agreeable", "nice",
    "awesome", "fabulous", "fantastic", "glorious", "marvelous", "splendid", "superb", "terrific", "admirable", "commendable",
    "noble", "excellent", "great", "incredible", "lively", "lovely", "magnificent", "outstanding", "peaceful",
    "kind", "rejoicing", "serene", "soothing", "supportive", "sympathetic", "tender", "vibrant", "warmhearted", "winsome"

    # Negative Synonyms
    "harsh", "sad", "angry", "upset", "depressed", "bitter", "gloomy", "anxious", "worried", "hostile", "resentful",
    "unhappy",  "irritable", "moody", "pessimistic", "fearful", "dismal", "horrible",  "awful", "nasty", "unpleasant",
    "terrible", "mean", "cruel",  "hurtful", "jealous", "malicious", "miserable", "regretful",  "scornful",
    "troubled","spiteful", "tense",  "vindictive", "vulgar", "wicked", "wretched", "abrasive", "agonizing",
    "evil", "brutal", "callous", "coldhearted", "disrespectful", "frustrated", "hateful", "hostile", "intolerant", "nervous",
    "repulsive"
 ]
\end{lstlisting}
\end{minipage}

\section{Embedding and Downsampling Ablations}
\label{appendix-downsampling}

\begin{figure}[h!]
    \centering
    \includegraphics[width=0.5\textwidth]{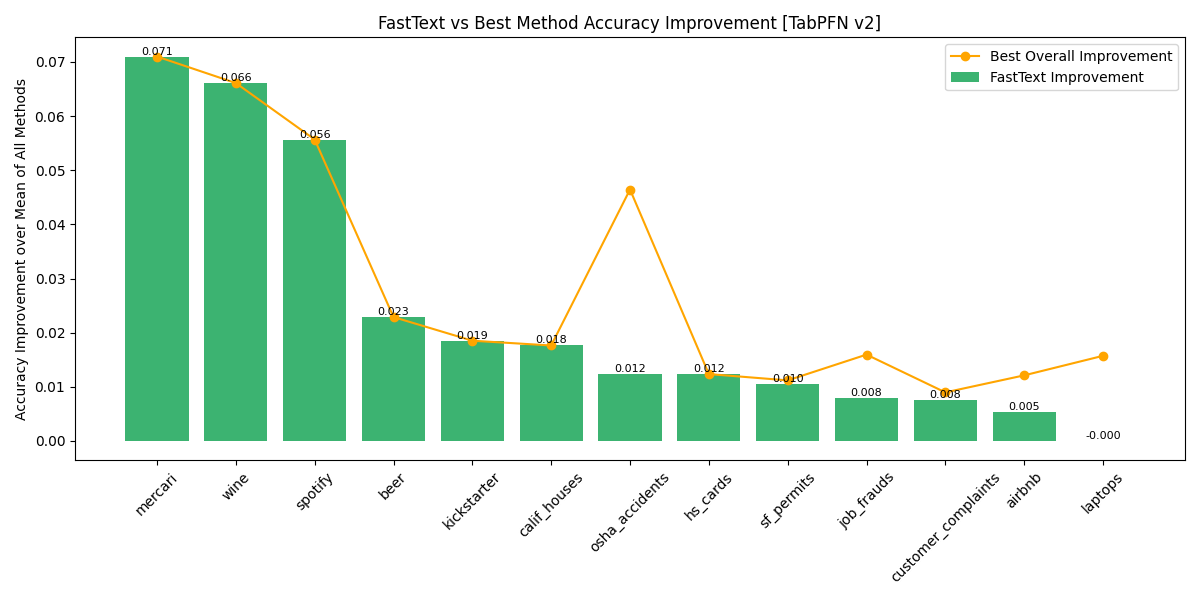}
    \caption{\textbf{Performance improvement with FastText over mean accuracy across all ablations on TabPFN v2}: FastText generally gives the best scores, with a few exceptions.These exceptions generally have longer length of texts than average.
    }
    \label{fig:fasttext_imp}
\end{figure}

\begin{figure}[h!]
    \centering
    \includegraphics[width=0.4\textwidth]{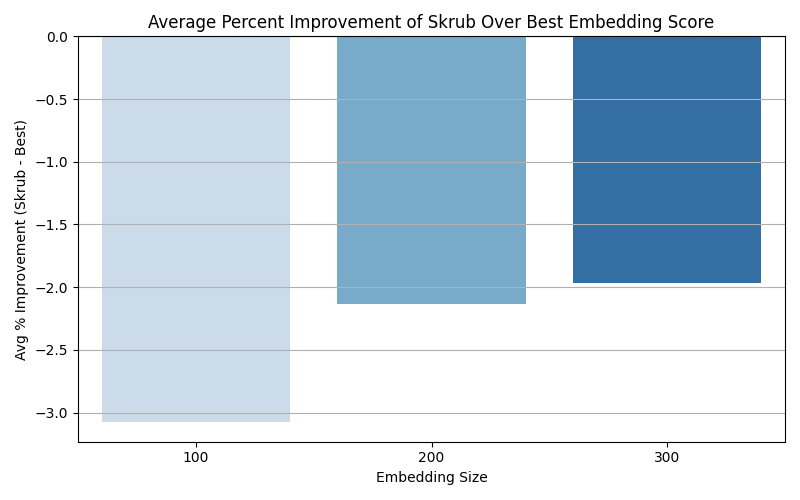}
    \caption{\textbf{Ablation over larger embedding sizes for Skrub GapEncoder over five datasets}: beer, calif\_houses, laptops, sf\_permits, wine. Bars show percentage difference between best score and best skrub embedding score (across all downsampling techniques), averaged across all chosen datasets. Skrub's performance improves with larger embedding dimensions, though the improvement slows down, and any further increase in dimensionality doesn't seem well motivated.}
    \label{fig:skrub}
\end{figure}

\begin{figure}[h!]
    \centering
    \begin{subfigure}[b]{0.45\textwidth}
        \centering
        \includegraphics[width=\linewidth]{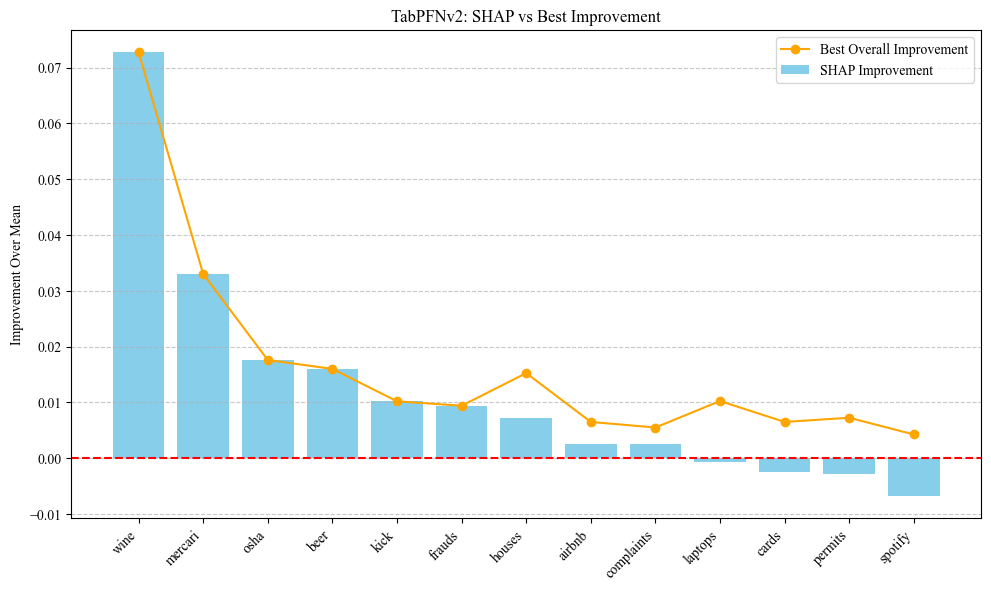}
        \caption{TabPFN v2 with SHAP}
        \label{fig:tabpfn_shap}
    \end{subfigure}
    \hfill
    \begin{subfigure}[b]{0.45\textwidth}
        \centering
        \includegraphics[width=\linewidth]{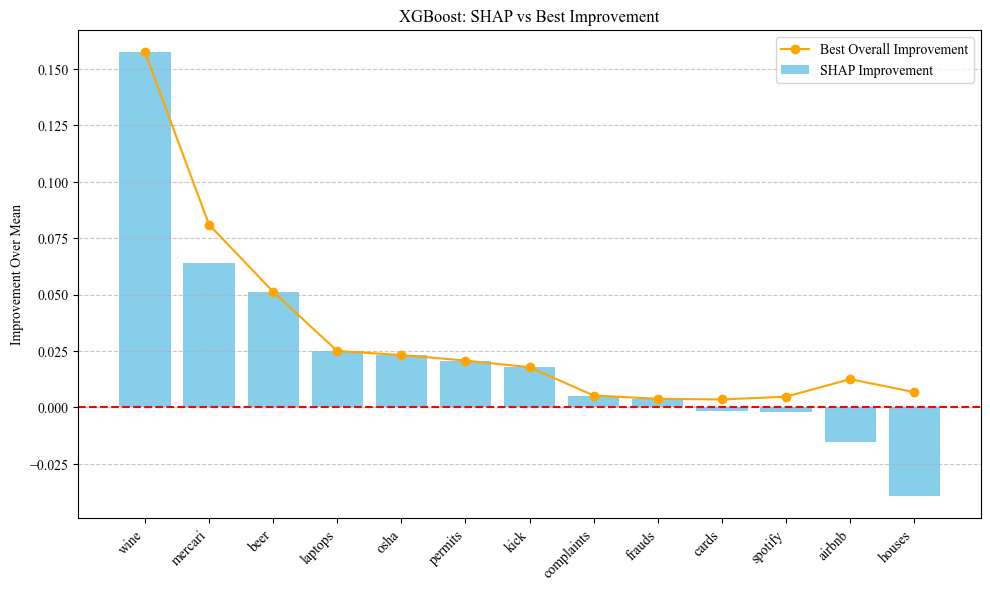}
        \caption{XGB with SHAP}
        \label{fig:xgb_shap}
    \end{subfigure}
    \caption{\textbf{SHAP performance improvement over mean accuracy across all ablations on FastText embeddings:} SHAP generally coincides with the best score using FastText for most datasets, on few exceptions, the performance is slightly below the best score.}
    \label{fig:shap_vs_best}
\end{figure}

\begin{figure}[h!]
    \centering
    \includegraphics[width=0.7\textwidth]{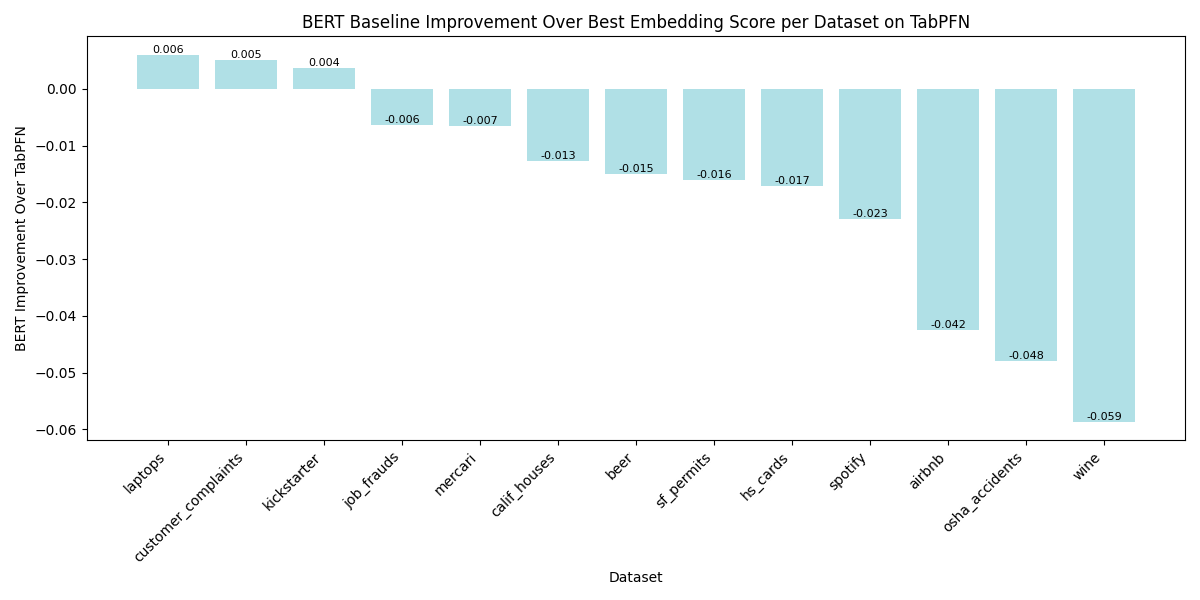}
    \caption{\textbf{Performance improvement with BERT variant over best accuracy across all ablations on TabPFN v2}: BERT is sub-optimal compared to best scores over other embeddings: FastText, Skrub and AGPIPE per dataset. This can be possibly due to higher embedding dimensions than the rest, hence downsampling drastically might have caused loss of information. }
    \label{fig:bert_ablations}
\end{figure}

We present our complete study of each embedding's performance under every downsampling technique. Tables \ref{tab:fasttext_comparison}, \ref{tab:skrub_comparison} and \ref{tab:ag_comparison} compare performance of TabPFN and XGBoost under different downsampling techniques for 3 embeddings FastText, Skrub and Autogluon feature generators respectively. The blanks generally mean the features were below the max-features threshold (300), or at times it's task specific, for e.g: T-Test and Anova are classification specific downsampling techniques so these techniques are skipped for regression tasks. 

\textbf{Comparison of FastText with Mean accuracy across all embeddings.} Figure \ref{fig:fasttext_imp} demonstrates improvement of FastText combined with the best performing downsampling technique over mean performance of all ablations for particular dataset. Here, it is clear FastText generally is the best performing embedding technique with a few exceptions.

\textbf{Comparison of SHAP with Mean accuracy across all downsampling techniques.} Figure \ref{fig:shap_vs_best} shows improvement of SHAP downsampling technique over mean accuracy across all downsampling techniques over FastText embeddings. SHAP generally achieves the highest score, with a few exceptions, although the magnitude of difference from the best score on these exceptions are low.

\textbf{Extended Ablations on higher dimensional embeddings using Skrub.} Since default configuration of Skrub's TableVectoriser uses GapEncoder with only 30 n-components, we include an extended ablation (see Figure \ref{fig:skrub}) for Skrub with higher embedding sizes to compare with the best score per dataset as in Table \ref{tab:main_table}. We observe that Skrub generally fails to outperform the best score on average across 5 datasets (beer,calif houses, laptops, sf permits and wine) from the pool, although the gap minimizes with higher embedding sizes. 

\textbf{Extended Ablations using BERT.} We additionally tested BERT embeddings on our dataset pool across all downsampling techniques. The difference between average BERT scores with best scores per dataset from \ref{tab:main_table} is in Figure \ref{fig:bert_ablations}. Complete ablation per embedding is shown in Table \ref{tab:bert_comparison}. 

Surprisingly, our experiments show that BERT embeddings do not generally outperform our baseline methods. This contrasts with prior work~\citep{grinsztajn2023vectorizingstringentriesdata}, which finds LLM embeddings more effective for rich textual features. Given that our datasets contain high-quality, well-structured text fields, the underperformance may be due to our substantial downsampling of BERT embeddings in our setup, imposed by hardware constraints. This opens up interesting directions for further investigation.

\begin{table}[!htbp]
\centering
\caption{Comparison of BERT using various feature selection techniques on TabPFN v2. \textbf{Bolded} values indicate the best performing method per dataset. Standard deviations (±) are shown as subscripts.}
\vskip 0.03in
\renewcommand{\arraystretch}{1.3}
\small
\resizebox{0.95\textwidth}{!}{
\begin{tabular}{l|*{8}{c}}
\toprule
\textbf{Dataset} & \textbf{t-test} & \textbf{anova} & \textbf{variance} & \textbf{pca} & \textbf{corr.} & \textbf{shap} & \textbf{rand.} & \textbf{lasso} \\
\midrule
airbnb & -- & -- & 0.641\textsubscript{\scriptsize±0.060} & 0.640\textsubscript{\scriptsize±0.063} & \textbf{0.650} & 0.648\textsubscript{\scriptsize±0.061} & 0.643 & 0.639\textsubscript{\scriptsize±0.060} \\
beer & -- & -- & 0.614\textsubscript{\scriptsize±0.013} & 0.624\textsubscript{\scriptsize±0.011} & 0.597 & \textbf{0.631\textsubscript{\scriptsize±0.008}} & 0.621 & 0.628\textsubscript{\scriptsize±0.010} \\
houses & -- & -- & 0.733\textsubscript{\scriptsize±0.057} & 0.736\textsubscript{\scriptsize±0.053} & 0.732 & \textbf{0.748\textsubscript{\scriptsize±0.053}} & 0.732 & 0.738\textsubscript{\scriptsize±0.054} \\
laptops & -- & -- & 0.896\textsubscript{\scriptsize±0.016} & 0.900\textsubscript{\scriptsize±0.021} & 0.903 & \textbf{0.914\textsubscript{\scriptsize±0.020}} & 0.905 & 0.913\textsubscript{\scriptsize±0.025} \\
mercari & -- & -- & 0.184\textsubscript{\scriptsize±0.032} & 0.170\textsubscript{\scriptsize±0.032} & \textbf{0.230} & 0.223\textsubscript{\scriptsize±0.032} & 0.185 & 0.181\textsubscript{\scriptsize±0.024} \\
permits & -- & -- & 0.484\textsubscript{\scriptsize±0.063} & 0.486\textsubscript{\scriptsize±0.061} & \textbf{0.489} & 0.476\textsubscript{\scriptsize±0.059} & 0.488 & 0.482\textsubscript{\scriptsize±0.059} \\
wine & -- & -- & 0.454\textsubscript{\scriptsize±0.149} & 0.455\textsubscript{\scriptsize±0.156} & 0.481 & \textbf{0.512\textsubscript{\scriptsize±0.132}} & 0.479 & 0.473\textsubscript{\scriptsize±0.141} \\
complaints & -- & \textbf{0.697\textsubscript{\scriptsize±0.005}} & 0.689\textsubscript{\scriptsize±0.008} & 0.692\textsubscript{\scriptsize±0.008} & -- & 0.689\textsubscript{\scriptsize±0.007} & 0.688 & 0.687\textsubscript{\scriptsize±0.010} \\
frauds & 0.945\textsubscript{\scriptsize±0.007} & 0.947\textsubscript{\scriptsize±0.007} & 0.948\textsubscript{\scriptsize±0.008} & 0.949\textsubscript{\scriptsize±0.009} & -- & \textbf{0.955\textsubscript{\scriptsize±0.008}} & 0.949 & 0.954\textsubscript{\scriptsize±0.006} \\
cards & -- & 0.683\textsubscript{\scriptsize±0.011} & 0.677\textsubscript{\scriptsize±0.010} & 0.673\textsubscript{\scriptsize±0.009} & -- & \textbf{0.695\textsubscript{\scriptsize±0.011}} & 0.683 & 0.690\textsubscript{\scriptsize±0.009} \\
kickstarter & 0.772\textsubscript{\scriptsize±0.010} & 0.779\textsubscript{\scriptsize±0.013} & 0.777\textsubscript{\scriptsize±0.010} & 0.776\textsubscript{\scriptsize±0.010} & -- & 0.780\textsubscript{\scriptsize±0.012} & \textbf{0.783} & 0.775\textsubscript{\scriptsize±0.010} \\
accidents & 0.550\textsubscript{\scriptsize±0.036} & 0.561\textsubscript{\scriptsize±0.037} & 0.523\textsubscript{\scriptsize±0.031} & 0.529\textsubscript{\scriptsize±0.031} & -- & \textbf{0.565\textsubscript{\scriptsize±0.035}} & 0.533 & 0.528\textsubscript{\scriptsize±0.030} \\
spotify & -- & 0.796\textsubscript{\scriptsize±0.027} & 0.803\textsubscript{\scriptsize±0.023} & \textbf{0.803\textsubscript{\scriptsize±0.023}} & -- & 0.799\textsubscript{\scriptsize±0.023} & 0.796 & 0.801\textsubscript{\scriptsize±0.021} \\
\bottomrule
\end{tabular}
}
\label{tab:bert_comparison}
\end{table}

\begin{table*}[!htbp]
\centering
\caption{Comparison of TabPFNv2 and XGBoost on FastText data embedding across downsampling techniques. \textbf{Bolded} values indicate the best performing method per model. Standard deviations (±) are shown as subscripts.}
\vskip 0.05in
\renewcommand{\arraystretch}{1.5}
\resizebox{\textwidth}{!}{
\begin{tabular}{l|*{8}{c}||*{8}{c}}
\toprule
\textbf{Dataset} & \multicolumn{8}{c||}{\textbf{TabPFNv2}} & \multicolumn{8}{c}{\textbf{XGBoost}} \\
 & \textbf{t-test} & \textbf{anova} & \textbf{variance} & \textbf{pca} & \textbf{corr.} & \textbf{shap} & \textbf{rand.} & \textbf{all} & \textbf{t-test} & \textbf{anova} & \textbf{variance} & \textbf{pca} & \textbf{corr.} & \textbf{shap} & \textbf{rand.} & \textbf{all} \\
\midrule
airbnb & -- & -- & 0.675\textsubscript{\scriptsize±0.043} & 0.675\textsubscript{\scriptsize±0.044} & \textbf{0.686\textsubscript{\scriptsize±0.050}} & 0.682\textsubscript{\scriptsize±0.054} & 0.682\textsubscript{\scriptsize±0.049} & -- & -- & -- & 0.582\textsubscript{\scriptsize±0.057} & \textbf{0.595\textsubscript{\scriptsize±0.036}} & 0.570\textsubscript{\scriptsize±0.037} & 0.567\textsubscript{\scriptsize±0.053} & 0.583\textsubscript{\scriptsize±0.045} & -- \\
beer & -- & -- & 0.632\textsubscript{\scriptsize±0.013} & 0.640\textsubscript{\scriptsize±0.015} & 0.617\textsubscript{\scriptsize±0.018} & \textbf{0.646\textsubscript{\scriptsize±0.023}} & 0.631\textsubscript{\scriptsize±0.021} & -- & -- & -- & 0.519\textsubscript{\scriptsize±0.024} & 0.520\textsubscript{\scriptsize±0.024} & 0.570\textsubscript{\scriptsize±0.020} & \textbf{0.594\textsubscript{\scriptsize±0.036}} & 0.562\textsubscript{\scriptsize±0.023} & -- \\
houses & -- & -- & 0.736\textsubscript{\scriptsize±0.040} & 0.741\textsubscript{\scriptsize±0.045} & \textbf{0.761\textsubscript{\scriptsize±0.039}} & 0.753\textsubscript{\scriptsize±0.035} & 0.745\textsubscript{\scriptsize±0.042} & -- & -- & -- & 0.702\textsubscript{\scriptsize±0.035} & 0.683\textsubscript{\scriptsize±0.039} & \textbf{0.706\textsubscript{\scriptsize±0.032}} & 0.660\textsubscript{\scriptsize±0.074} & 0.706\textsubscript{\scriptsize±0.032} & -- \\
laptops & -- & -- & 0.880\textsubscript{\scriptsize±0.016} & 0.874\textsubscript{\scriptsize±0.024} & 0.884\textsubscript{\scriptsize±0.009} & 0.882\textsubscript{\scriptsize±0.014} & \textbf{0.893\textsubscript{\scriptsize±0.017}} & -- & -- & -- & 0.761\textsubscript{\scriptsize±0.033} & 0.770\textsubscript{\scriptsize±0.031} & 0.806\textsubscript{\scriptsize±0.048} & \textbf{0.811\textsubscript{\scriptsize±0.037}} & 0.807\textsubscript{\scriptsize±0.037} & -- \\
mercari & -- & -- & 0.199\textsubscript{\scriptsize±0.050} & 0.197\textsubscript{\scriptsize±0.041} & 0.221\textsubscript{\scriptsize±0.036} & \textbf{0.237\textsubscript{\scriptsize±0.050}} & 0.199\textsubscript{\scriptsize±0.042} & -- & -- & -- & -0.006\textsubscript{\scriptsize±0.194} & -0.014\textsubscript{\scriptsize±0.227} & \textbf{0.127\textsubscript{\scriptsize±0.046}} & 0.110\textsubscript{\scriptsize±0.062} & 0.077\textsubscript{\scriptsize±0.125} & -- \\
permits & -- & -- & \textbf{0.504\textsubscript{\scriptsize±0.058}} & 0.500\textsubscript{\scriptsize±0.058} & 0.489\textsubscript{\scriptsize±0.066} & 0.494\textsubscript{\scriptsize±0.062} & 0.494\textsubscript{\scriptsize±0.062} & -- & -- & -- & 0.406\textsubscript{\scriptsize±0.041} & 0.400\textsubscript{\scriptsize±0.052} & 0.412\textsubscript{\scriptsize±0.034} & \textbf{0.426\textsubscript{\scriptsize±0.040}} & 0.403\textsubscript{\scriptsize±0.061} & -- \\
wine & -- & -- & 0.486\textsubscript{\scriptsize±0.110} & 0.479\textsubscript{\scriptsize±0.113} & 0.512\textsubscript{\scriptsize±0.115} & \textbf{0.571\textsubscript{\scriptsize±0.115}} & 0.516\textsubscript{\scriptsize±0.093} & -- & -- & -- & 0.149\textsubscript{\scriptsize±0.237} & 0.153\textsubscript{\scriptsize±0.223} & 0.343\textsubscript{\scriptsize±0.177} & \textbf{0.394\textsubscript{\scriptsize±0.098}} & 0.301\textsubscript{\scriptsize±0.202} & -- \\
\midrule
complaints & -- & \textbf{0.691\textsubscript{\scriptsize±0.025}} & 0.681\textsubscript{\scriptsize±0.027} & 0.683\textsubscript{\scriptsize±0.029} & -- & 0.688\textsubscript{\scriptsize±0.024} & 0.687\textsubscript{\scriptsize±0.027} & -- & -- & 0.666\textsubscript{\scriptsize±0.021} & 0.667\textsubscript{\scriptsize±0.026} & 0.662\textsubscript{\scriptsize±0.029} & -- & \textbf{0.672\textsubscript{\scriptsize±0.021}} & 0.672\textsubscript{\scriptsize±0.029} & -- \\
frauds & 0.945\textsubscript{\scriptsize±0.005} & 0.949\textsubscript{\scriptsize±0.005} & 0.941\textsubscript{\scriptsize±0.014} & 0.937\textsubscript{\scriptsize±0.012} & -- & \textbf{0.954\textsubscript{\scriptsize±0.005}} & 0.951\textsubscript{\scriptsize±0.007} & -- & 0.947\textsubscript{\scriptsize±0.001} & 0.948\textsubscript{\scriptsize±0.005} & 0.944\textsubscript{\scriptsize±0.012} & 0.939\textsubscript{\scriptsize±0.010} & -- & \textbf{0.949\textsubscript{\scriptsize±0.005}} & 0.948\textsubscript{\scriptsize±0.007} & -- \\
cards & -- & 0.709\textsubscript{\scriptsize±0.011} & 0.707\textsubscript{\scriptsize±0.008} & \textbf{0.712\textsubscript{\scriptsize±0.009}} & -- & 0.703\textsubscript{\scriptsize±0.008} & 0.694\textsubscript{\scriptsize±0.008} & -- & -- & 0.726\textsubscript{\scriptsize±0.005} & \textbf{0.729\textsubscript{\scriptsize±0.010}} & 0.729\textsubscript{\scriptsize±0.012} & -- & 0.724\textsubscript{\scriptsize±0.011} & 0.718\textsubscript{\scriptsize±0.008} & -- \\
kick & 0.772\textsubscript{\scriptsize±0.009} & 0.772\textsubscript{\scriptsize±0.013} & 0.764\textsubscript{\scriptsize±0.010} & 0.766\textsubscript{\scriptsize±0.011} & -- & \textbf{0.779\textsubscript{\scriptsize±0.016}} & 0.770\textsubscript{\scriptsize±0.015} & -- & 0.755\textsubscript{\scriptsize±0.007} & 0.755\textsubscript{\scriptsize±0.012} & 0.749\textsubscript{\scriptsize±0.016} & 0.750\textsubscript{\scriptsize±0.023} & -- & \textbf{0.769\textsubscript{\scriptsize±0.014}} & 0.747\textsubscript{\scriptsize±0.011} & -- \\
osha & 0.575\textsubscript{\scriptsize±0.008} & 0.566\textsubscript{\scriptsize±0.009} & 0.551\textsubscript{\scriptsize±0.023} & 0.563\textsubscript{\scriptsize±0.019} & -- & \textbf{0.579\textsubscript{\scriptsize±0.015}} & 0.552\textsubscript{\scriptsize±0.018} & -- & 0.561\textsubscript{\scriptsize±0.029} & 0.551\textsubscript{\scriptsize±0.016} & 0.538\textsubscript{\scriptsize±0.006} & 0.535\textsubscript{\scriptsize±0.018} & -- & \textbf{0.571\textsubscript{\scriptsize±0.009}} & 0.554\textsubscript{\scriptsize±0.017} & -- \\
spotify & -- & 0.822\textsubscript{\scriptsize±0.013} & 0.823\textsubscript{\scriptsize±0.016} & \textbf{0.826\textsubscript{\scriptsize±0.014}} & -- & 0.815\textsubscript{\scriptsize±0.010} & 0.816\textsubscript{\scriptsize±0.012} & -- & -- & \textbf{0.814\textsubscript{\scriptsize±0.007}} & 0.812\textsubscript{\scriptsize±0.009} & 0.811\textsubscript{\scriptsize±0.009} & -- & 0.807\textsubscript{\scriptsize±0.012} & 0.800\textsubscript{\scriptsize±0.010} & -- \\
\bottomrule
\end{tabular}
}

\label{tab:fasttext_comparison}
\end{table*}

\begin{table*}[!htbp]
\centering
\caption{Comparison of TabPFNv2 and XGBoost using Skrub features across downsampling techniques. \textbf{Bolded} values indicate the best performing method per model. Standard deviations (±) are shown as subscripts.}
\vskip 0.05in
\renewcommand{\arraystretch}{1.9}
\resizebox{\textwidth}{!}{
\begin{tabular}{l|*{8}{c}||*{8}{c}}
\toprule
\textbf{Dataset} & \multicolumn{8}{c||}{\textbf{TabPFNv2}} & \multicolumn{8}{c}{\textbf{XGBoost}} \\
 & \textbf{t-test} & \textbf{anova} & \textbf{variance} & \textbf{pca} & \textbf{corr.} & \textbf{shap} & \textbf{rand.} & \textbf{all} & \textbf{t-test} & \textbf{anova} & \textbf{variance} & \textbf{pca} & \textbf{corr.} & \textbf{shap} & \textbf{rand.} & \textbf{all} \\
\midrule
airbnb & -- & -- & 0.680\textsubscript{\scriptsize±0.047} & 0.680\textsubscript{\scriptsize±0.049} & 0.679\textsubscript{\scriptsize±0.049} & 0.681\textsubscript{\scriptsize±0.048} & \textbf{0.681\textsubscript{\scriptsize±0.047}} & -- & -- & -- & 0.552\textsubscript{\scriptsize±0.027} & 0.541\textsubscript{\scriptsize±0.047} & \textbf{0.568\textsubscript{\scriptsize±0.043}} & 0.562\textsubscript{\scriptsize±0.047} & 0.545\textsubscript{\scriptsize±0.059} & -- \\
beer & -- & -- & -- & -- & -- & -- & -- & \textbf{0.634\textsubscript{\scriptsize±0.005}} & -- & -- & -- & -- & -- & -- & -- & \textbf{0.560\textsubscript{\scriptsize±0.021}} \\
houses & -- & -- & -- & -- & -- & -- & -- & \textbf{0.738\textsubscript{\scriptsize±0.037}} & -- & -- & -- & -- & -- & -- & -- & \textbf{0.689\textsubscript{\scriptsize±0.012}} \\
laptops & -- & -- & 0.895\textsubscript{\scriptsize±0.025} & 0.896\textsubscript{\scriptsize±0.023} & \textbf{0.908\textsubscript{\scriptsize±0.022}} & 0.900\textsubscript{\scriptsize±0.022} & 0.896\textsubscript{\scriptsize±0.024} & -- & -- & -- & 0.809\textsubscript{\scriptsize±0.044} & 0.825\textsubscript{\scriptsize±0.040} & 0.829\textsubscript{\scriptsize±0.053} & \textbf{0.833\textsubscript{\scriptsize±0.046}} & 0.822\textsubscript{\scriptsize±0.041} & -- \\
mercari & -- & -- & -- & -- & -- & -- & -- & \textbf{0.155\textsubscript{\scriptsize±0.028}} & -- & -- & -- & -- & -- & -- & -- & \textbf{0.099\textsubscript{\scriptsize±0.046} }\\
permits & -- & -- & -- & -- & -- & -- & -- & \textbf{0.505\textsubscript{\scriptsize±0.060}} & -- & -- & -- & -- & -- & -- & -- & \textbf{0.437\textsubscript{\scriptsize±0.053}} \\
wine & -- & -- & -- & -- & -- & -- & -- & \textbf{0.467\textsubscript{\scriptsize±0.145}} & -- & -- & -- & -- & -- & -- & -- & \textbf{0.213\textsubscript{\scriptsize±0.263}} \\
\midrule
complaints & -- & -- & -- & -- & -- & -- & -- & \textbf{0.692\textsubscript{\scriptsize±0.027}} & -- & -- & -- & -- & -- & -- & -- & \textbf{0.678\textsubscript{\scriptsize±0.037}} \\
frauds & -- & -- & -- & -- & -- & -- & -- & \textbf{0.953\textsubscript{\scriptsize±0.007}} & -- & -- & -- & -- & -- & -- & -- & \textbf{0.952\textsubscript{\scriptsize±0.007}} \\
cards & -- & -- & -- & -- & -- & -- & -- & \textbf{0.689\textsubscript{\scriptsize±0.011}} & -- & -- & -- & -- & -- & -- & -- & \textbf{0.723\textsubscript{\scriptsize±0.013}} \\
kick & -- & -- & -- & -- & -- & -- & -- & \textbf{0.763\textsubscript{\scriptsize±0.019}} & -- & -- & -- & -- & -- & -- & -- & \textbf{0.747\textsubscript{\scriptsize±0.023}} \\
osha & -- & -- & -- & -- & -- & -- & -- & \textbf{0.560\textsubscript{\scriptsize±0.015}} & -- & -- & -- & -- & -- & -- & -- & \textbf{0.535\textsubscript{\scriptsize±0.018}} \\
spotify & -- & -- & -- & -- & -- & -- & -- & \textbf{0.810\textsubscript{\scriptsize±0.007}} & -- & -- & -- & -- & -- & -- & -- & \textbf{0.760\textsubscript{\scriptsize±0.021}}\\
\bottomrule
\end{tabular}
}
\label{tab:skrub_comparison}
\end{table*}

\begin{table*}[!htbp]
\centering
\caption{Comparison of TabPFNv2 and XGBoost using AutoMLPipelineFeatureGenerator (AGPIPE) across various downsampling techniques. \textbf{Bolded} values indicate the best performing method per model. Standard deviations (±) are shown as subscripts.}
\vskip 0.05in
\renewcommand{\arraystretch}{1.9}
\resizebox{\textwidth}{!}{
\begin{tabular}{l|*{8}{c}||*{8}{c}}
\toprule
\textbf{Dataset} & \multicolumn{8}{c||}{\textbf{TabPFNv2}} & \multicolumn{8}{c}{\textbf{XGBoost}} \\
 & \textbf{t-test} & \textbf{anova} & \textbf{variance} & \textbf{pca} & \textbf{corr.} & \textbf{shap} & \textbf{rand.} & \textbf{all} & \textbf{t-test} & \textbf{anova} & \textbf{variance} & \textbf{pca} & \textbf{corr.} & \textbf{shap} & \textbf{rand.} & \textbf{all} \\
\midrule
airbnb & -- & -- & 0.673\textsubscript{\scriptsize±0.048} & 0.675\textsubscript{\scriptsize±0.044} & 0.684\textsubscript{\scriptsize±0.047} & \textbf{0.692\textsubscript{\scriptsize±0.048}} & 0.678\textsubscript{\scriptsize±0.047} & -- & -- & -- & 0.603\textsubscript{\scriptsize±0.043} & 0.601\textsubscript{\scriptsize±0.034} & \textbf{0.617\textsubscript{\scriptsize±0.057}} & 0.603\textsubscript{\scriptsize±0.060} & 0.598\textsubscript{\scriptsize±0.040} & -- \\
beer & -- & -- & 0.609\textsubscript{\scriptsize±0.016} & 0.612\textsubscript{\scriptsize±0.015} & 0.609\textsubscript{\scriptsize±0.019} & \textbf{0.618\textsubscript{\scriptsize±0.014}} & 0.603\textsubscript{\scriptsize±0.020} & -- & -- & -- & \textbf{0.559\textsubscript{\scriptsize±0.010}} & 0.540\textsubscript{\scriptsize±0.023} & 0.548\textsubscript{\scriptsize±0.024} & 0.553\textsubscript{\scriptsize±0.030} & 0.539\textsubscript{\scriptsize±0.023} & -- \\
houses & -- & -- & 0.736\textsubscript{\scriptsize±0.037} & 0.738\textsubscript{\scriptsize±0.030} & 0.741\textsubscript{\scriptsize±0.032} & \textbf{0.747\textsubscript{\scriptsize±0.039}} & 0.735\textsubscript{\scriptsize±0.026} & -- & -- & -- & 0.678\textsubscript{\scriptsize±0.044} & 0.687\textsubscript{\scriptsize±0.041} & \textbf{0.707\textsubscript{\scriptsize±0.033}} & 0.696\textsubscript{\scriptsize±0.020} & 0.678\textsubscript{\scriptsize±0.027} & -- \\
laptops & -- & -- & 0.897\textsubscript{\scriptsize±0.027} & 0.899\textsubscript{\scriptsize±0.029} & 0.890\textsubscript{\scriptsize±0.021} & \textbf{0.902\textsubscript{\scriptsize±0.023}} & 0.895\textsubscript{\scriptsize±0.018} & -- & -- & -- & 0.804\textsubscript{\scriptsize±0.041} & 0.799\textsubscript{\scriptsize±0.047} & \textbf{0.822\textsubscript{\scriptsize±0.016}} & 0.815\textsubscript{\scriptsize±0.042} & 0.804\textsubscript{\scriptsize±0.037} & -- \\
mercari & -- & -- & 0.119\textsubscript{\scriptsize±0.021} & 0.123\textsubscript{\scriptsize±0.023} & 0.151\textsubscript{\scriptsize±0.022} & \textbf{0.173\textsubscript{\scriptsize±0.037}} & 0.051\textsubscript{\scriptsize±0.021} & -- & -- & -- & 0.096\textsubscript{\scriptsize±0.092} & 0.072\textsubscript{\scriptsize±0.042} & 0.096\textsubscript{\scriptsize±0.192} & \textbf{0.096\textsubscript{\scriptsize±0.070}} & -0.266\textsubscript{\scriptsize±0.270} & -- \\
permits & -- & -- & 0.491\textsubscript{\scriptsize±0.061} & 0.482\textsubscript{\scriptsize±0.061} & \textbf{0.494\textsubscript{\scriptsize±0.065}} & 0.492\textsubscript{\scriptsize±0.058} & 0.488\textsubscript{\scriptsize±0.052} & -- & -- & -- & 0.437\textsubscript{\scriptsize±0.046} & 0.437\textsubscript{\scriptsize±0.025} & \textbf{0.451\textsubscript{\scriptsize±0.050}} & 0.420\textsubscript{\scriptsize±0.054} & 0.436\textsubscript{\scriptsize±0.032} & -- \\
wine & -- & -- & 0.505\textsubscript{\scriptsize±0.124} & 0.503\textsubscript{\scriptsize±0.114} & \textbf{0.531\textsubscript{\scriptsize±0.110}} & 0.527\textsubscript{\scriptsize±0.135} & 0.456\textsubscript{\scriptsize±0.134} & -- & -- & -- & 0.283\textsubscript{\scriptsize±0.305} & 0.281\textsubscript{\scriptsize±0.288} & 0.317\textsubscript{\scriptsize±0.354} & \textbf{0.339\textsubscript{\scriptsize±0.243}} & 0.184\textsubscript{\scriptsize±0.302} & -- \\
\midrule
complaints & -- & 0.673\textsubscript{\scriptsize±0.029} & 0.681\textsubscript{\scriptsize±0.034} & \textbf{0.682\textsubscript{\scriptsize±0.027}} & -- & -- & 0.677\textsubscript{\scriptsize±0.030} & 0.678\textsubscript{\scriptsize±0.029} & -- & 0.666\textsubscript{\scriptsize±0.021} & \textbf{0.669\textsubscript{\scriptsize±0.018}} & 0.650\textsubscript{\scriptsize±0.014} & -- & 0.664\textsubscript{\scriptsize±0.019} & 0.662\textsubscript{\scriptsize±0.020} & -- \\
frauds & 0.950\textsubscript{\scriptsize±0.006} & 0.947\textsubscript{\scriptsize±0.007} & 0.945\textsubscript{\scriptsize±0.006} & 0.946\textsubscript{\scriptsize±0.005} & -- & \textbf{0.962\textsubscript{\scriptsize±0.008}} & 0.915\textsubscript{\scriptsize±0.007} & -- & 0.940\textsubscript{\scriptsize±0.006} & 0.946\textsubscript{\scriptsize±0.002} & 0.953\textsubscript{\scriptsize±0.005} & 0.952\textsubscript{\scriptsize±0.005} & -- & \textbf{0.958\textsubscript{\scriptsize±0.004}} & 0.919\textsubscript{\scriptsize±0.008} & -- \\
cards & -- & -- & -- & -- & -- & -- & -- & \textbf{0.683\textsubscript{\scriptsize±0.010}} & -- & -- & -- & -- & -- & -- & -- & \textbf{0.698\textsubscript{\scriptsize±0.015}} \\
kick & 0.756\textsubscript{\scriptsize±0.021} & 0.700\textsubscript{\scriptsize±0.017} & 0.762\textsubscript{\scriptsize±0.015} & 0.763\textsubscript{\scriptsize±0.018} & -- & \textbf{0.767\textsubscript{\scriptsize±0.015}} & 0.757\textsubscript{\scriptsize±0.017} & -- & 0.721\textsubscript{\scriptsize±0.020} & 0.679\textsubscript{\scriptsize±0.014} & 0.721\textsubscript{\scriptsize±0.008} & 0.721\textsubscript{\scriptsize±0.012} & -- & \textbf{0.731\textsubscript{\scriptsize±0.010}} & 0.729\textsubscript{\scriptsize±0.020} & -- \\
osha & 0.568\textsubscript{\scriptsize±0.012} & 0.556\textsubscript{\scriptsize±0.011} & 0.560\textsubscript{\scriptsize±0.018} & 0.565\textsubscript{\scriptsize±0.014} & -- & \textbf{0.613\textsubscript{\scriptsize±0.006}} & 0.557\textsubscript{\scriptsize±0.013} & -- & 0.557\textsubscript{\scriptsize±0.011} & 0.535\textsubscript{\scriptsize±0.013} & 0.564\textsubscript{\scriptsize±0.015} & 0.559\textsubscript{\scriptsize±0.025} & -- & \textbf{0.599\textsubscript{\scriptsize±0.020}} & 0.542\textsubscript{\scriptsize±0.007} & -- \\
spotify & -- & 0.713\textsubscript{\scriptsize±0.011} & \textbf{0.718\textsubscript{\scriptsize±0.006}} & 0.713\textsubscript{\scriptsize±0.007} & -- & 0.716\textsubscript{\scriptsize±0.011} & 0.704\textsubscript{\scriptsize±0.015} & -- & -- & 0.712\textsubscript{\scriptsize±0.028} & \textbf{0.713\textsubscript{\scriptsize±0.015}} & 0.711\textsubscript{\scriptsize±0.023} & -- & 0.706\textsubscript{\scriptsize±0.019} & 0.703\textsubscript{\scriptsize±0.023} & -- \\
\bottomrule
\end{tabular}
}

\label{tab:ag_comparison}
\end{table*}

\newpage
\section{Previous Benchmarks Accessment}
\label{appendix-other-benchmnark-datasets}

We provide a detailed review of two prior tabular benchmarks involving text features: CARTE \cite{kim2024cartepretrainingtransfertabular} and Amazon's  Multimodal AutoML for Tabular Data \cite{autogluon-text-tabular-benchmark}. Our primary focus is CARTE, as it is the more recent and comprehensive of the two.

The CARTE benchmark includes 40 regression and 11 classification datasets. However, upon close examination, we find that many of these datasets are ill-suited for benchmarking systems that aim to handle tabular data with semantically rich free-text features. Below, we highlight common issues and explain why a significant portion of the benchmark is incompatible with our goal of an evaluation design representative of real-world tabular data with text.

\subsection{TL;DR}
We review 51 datasets from the CARTE benchmark and 18 datasets from the "Multimodal AutoML for Tabular Data with Text Fields" paper. While both were instrumental in shaping our benchmark design, our detailed inspection revealed substantial limitations in their suitability for real-world evaluation of text-tabular systems.

For the 18 datasets, the primary issue was their strong bias toward text-heavy datasets, rendering most tasks better suited for language models than for general-purpose tabular learners. Nonetheless, we retained,  with some degree of modifications, six of their datasets that exhibited balanced multimodal features.

The CARTE benchmark, despite its breadth, suffers from domain redundancy, overuse of artificial targets, and limited semantic depth in its text columns. Many string features serve as identifiers or location codes, offering little benefit to textual encoders. Moreover, several datasets undergo significant undocumented preprocessing, making reproduction difficult.

Based on the discussed shortcomings, we defined a short list of lightweight selection rules to guide our manual dataset curation, as detailed in the main body of the paper. While defining these rules, we closely examined the limitations of prior benchmarks. To support and illustrate these decisions, the following section explores several examples in more detail. Some of the reasoning, such as assessing feature similarity across datasets or the quality of textual signals, requires a degree of human-like meta-understanding. To approximate this process in a transparent and relatively independent way, we incorporate outputs from a highly capable LLM (ChatGPT-4o via API), along with the corresponding prompts.

Ultimately, we used both benchmarks as a springboard, retaining or adapting a few datasets and collecting new ones to construct a more principled and representative benchmark. Our final selection prioritizes semantic textual inputs, domain diversity, and predictive alignment. 

\subsection{Domain Redundancy and Target Overlap}
One of the most immediate issues we observed is the reuse of the same dataset under different targets. For example, several wine review datasets are evaluated both on price and rating targets:

\begin{table}[h!]
\caption{List of datasets reused with different target columns.}
\centering
\begin{tabular}{lccc}
\toprule
\textbf{Dataset} & \textbf{Target 1} & \textbf{Target 2} & \textbf{Retains the Other Target?} \\
\midrule
Wine.com & Rating & Prices & Yes \\
Wine Enthusiasts & Points & Price & Yes \\
Wine Vivino & Price & Rating & Yes \\
\bottomrule
\end{tabular}
\label{table:dataset-targets}
\end{table}

From a modeling perspective, targets like price and rating are often strongly correlated and are unlikely to offer an independent benchmarking signal. Including both only serves to overweight particular domains and can bias results toward models that perform well in those areas.

\subsection{Limited Domain Diversity}

Although the CARTE benchmark includes over 50 datasets, many of them share the same prediction targets and come from overlapping domains. These two factors are often entangled, leading to much lower effective coverage than the “50+ dataset” label suggests.

When grouped by thematic content or feature-space similarity, the 51 datasets consolidate into approximately 14 to 20 distinct domains.
The exact number depends on how strictly one chooses to group datasets. In certain borderline cases, the topics may differ, but the feature schemas and prediction targets are highly similar, which provides a reasonable justification for merging.
For example, coffee and alcohol reviews exhibit nearly identical feature schemas and prediction targets, and can reasonably be grouped under a single ``consumables/sensory" domain.

To reduce manual bias in domain classification, we leverage large language models for an independent grouping. Specifically, we use OpenAI's GPT-4o to assess dataset similarity based solely on feature names, representative values, and a prompt describing the benchmark context. This LLM-driven process yields 14 coherent domain groupings, aligning with the lower bound of our manual estimate.

Below, we present our manually derived, more permissive domain categorization of datasets from the CARTE benchmark:

\vspace{0.5em}
\textbf{Regression Dataset Domains}
\vspace{-0.5em}
\begin{itemize}[itemsep=0.3em, topsep=0.3em]
  \item \textbf{Movies and Shows}: Anime Planet, Japanese Anime, K-Drama, Filmtv Movies, Mydramalist, Rotten Tomatoes
  \item \textbf{Kid Products}: Babies R Us, Buy Buy Baby
  \item \textbf{Alcohol Reviews}: Beer Rating, Wikiliq (Beer \& Spirits), Wina Poland, Wine.com, WineEnthusiasts, WineVivino
  \item \textbf{Bike \& Car Listings}: Bikewale, Bikedekho, Cardekho, Used Cars 24, Used Cars Benz Italy, UsedCars.com, Used Cars Pakistan, Used Cars Saudi Arabia
  \item \textbf{Salaries}: Employee remuneration, Employee Salaries, ML/DS Salaries
  \item \textbf{Academic Text}: Journal Score JCR, Journal Score SJR
  \item \textbf{Other}: Clear Corpus, Company Employees, Fifa22 Players, Museums, Prescription Drugs, Videogame Sales, US Accidents
\end{itemize}

\vspace{0.5em}
\textbf{Classification Dataset Domains}
\vspace{-0.5em}
\begin{itemize}[itemsep=0.3em, topsep=0.3em]
  \item \textbf{Restaurants}: Zomato, Yelp, Michelin
  \item \textbf{Cross-domain}: Whisky (Alcohol), Roger Ebert (Movies), US Accidents (also used in regression)
  \item \textbf{Miscellaneous}: NBA Draft, Spotify, Ramen rating, Chocolate Bars Ratings, Coffee Ratings
\end{itemize}
\subsection{Qualitative Evaluation via Semantic Matching}

To better understand the extent of feature overlap across datasets, we performed a semantic feature comparison using OpenAI's \textit{gpt-4o}. Given the cost of evaluating all pairwise combinations, we present a smaller pilot study using a standardized prompt to extract semantically aligned and misaligned features.

\begin{lstlisting}[language=Python, numbers=none, caption={Prompt for feature similarity decision via ChatGPT's 4o model. The $<$format template$>$ is excluded from the prompt to save space, its structure is visible from the example return in Listing \ref{lst:feature_sim_return}.}, label={lst:prompt_feature_sim}]
prompt_template = """
    Analyze and compare the feature space of two datasets: {dataset1_name} and {dataset2_name}.
    Identify relationships between feature names using semantic reasoning.
    
    **Your Task:**
    1. **Find Similar Features**: Identify columns that represent the same concept, even if their names differ.
       - Match based on **data type, structure, and naming conventions**, rather than relying solely on example values.
       - Consider cases where **one feature in a dataset maps to multiple features** in the other dataset.
       - Note: When a column represents an inherent property of the entity (such as its name, title, or composition/build/materials), treat it as similar across datasets unless context clearly indicates a different meaning.
    
    2. **Identify Dissimilar Features**: Columns that do not have a meaningful equivalent in the other dataset.
       - Consider **data type mismatches** (e.g., numeric vs. categorical).
       - Features that belong to completely different contexts should be classified as dissimilar.
       - For instance, even if the same term (e.g., "location") is used in both datasets, they should only be considered similar if their contexts align.
    
    ### Additional Guidelines:
    - **Return column names, with original example values.** Do not assume or generate example values.
    - Consider **semantic similarity** beyond direct string matching.
    - Account for **differences in feature naming conventions** (e.g., "price" vs. "cost", "region" vs. "province").
    - **Preserve structured output strictly in JSON format** - avoid any additional text or explanations.
    - Make sure you always return a pair of features for the "similar_features" section.
    
    ### Expected Output:
    {format_template}
    
    ### Dataset 1: {dataset1_name}
    {dataset1_values}
    
    ### Dataset 2: {dataset2_name}
    {dataset2_values}
"""
\end{lstlisting}

\begin{lstlisting}[language=Python, numbers=none, caption={Sample output: comparing coffee vs ramen datasets}, label={lst:feature_sim_return}]
"carte_ramen_ratings vs carte_coffee_ratings": {
    "similar_features": [
        {
            "dataset1_col_name": {"Brand": "MIT"},
            "dataset2_col_name": {"roaster": "A.R.C."},
            "reason": "Both represent the product's manufacturer or origin."
        },
        ...
    ],
    "dissimilar_features": {
        "dataset1": [{"col_name": "Style", ...}],
        "dataset2": [{"col_name": "origin", ...}]
    }
}
\end{lstlisting}

This structured approach highlights redundancy in several domains within the CARTE benchmark and reinforces the need for more careful dataset selection.

Broad domain diversity is important for evaluating how well models generalize across varying tasks and data distributions. When a benchmark heavily reuses a specific domain or target type, models may appear to perform well simply by overfitting to recurring patterns, rather than demonstrating true adaptability. This can lead to inflated performance estimates and misleading conclusions about general-purpose effectiveness.

\begin{figure*}[h!]
\centering
\includegraphics[width=\textwidth]{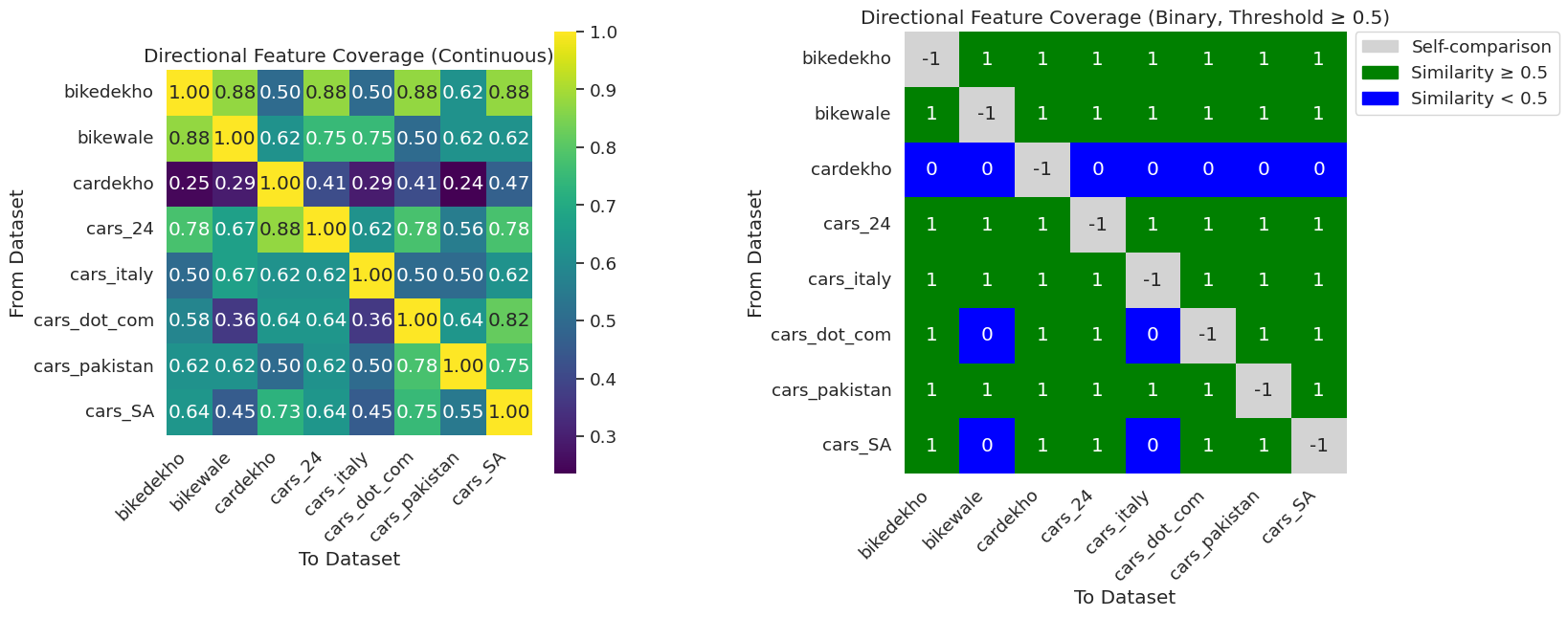}
\caption{Directional feature coverage between used vehicle listing datasets. Left: proportion of semantically similar features from dataset A to B. Right: binary thresholded version where coverage $\geq 0.5$ is highlighted in blue. Grey cells indicate missing comparisons.}
\label{similar_data-plot}
\end{figure*}

\begin{figure*}[h!]
\centering
\includegraphics[width=\textwidth]{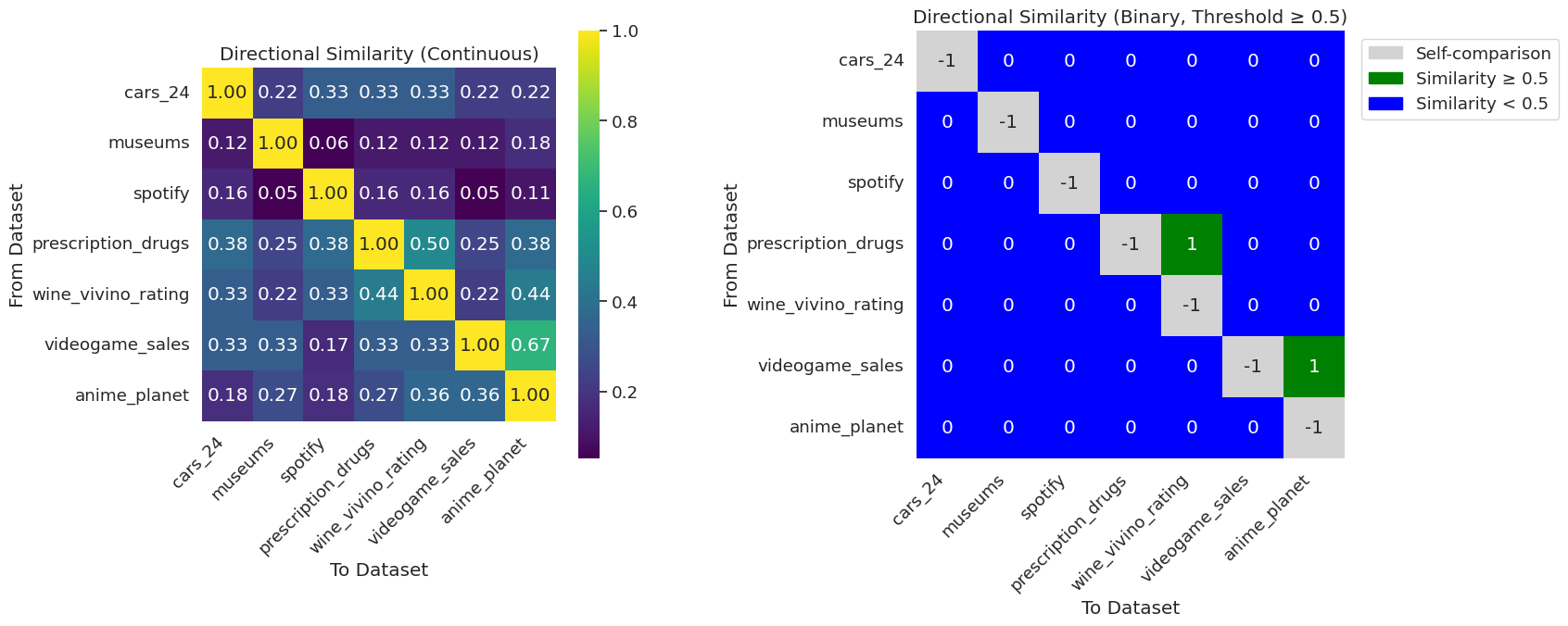}
\caption{Directional feature coverage between cross-domain datasets with low expected similarity. The left heatmap shows continuous coverage from dataset A to B; the right applies a binary threshold of $\geq 0.5$. Grey cells indicate unpaired or unmatched datasets.}
\label{dissimilar_data-plot}
\end{figure*}

\paragraph{Visualizing Feature Overlap.}
Figures~\ref{similar_data-plot} and~\ref{dissimilar_data-plot} visualize the directional feature coverage between dataset pairs using both continuous and binary heatmaps. Each matrix cell $(A \rightarrow B)$ quantifies the proportion of features in dataset $A$ that were semantically matched to features in dataset $B$, computed as $\#\text{similar features} / (\#\text{similar} + \#\text{dissimilar features in } A)$. This directional score captures asymmetries due to differing feature schema sizes.

The left heatmaps display continuous similarity scores, with masked cells where no semantic match was found. The right heatmaps binarize these scores using a $0.5$ threshold to indicate strong vs. weak directional similarity.
Blue cells indicate weak or no similarity, green cells indicate strong directional similarity ($\geq 0.5$), and light grey diagonal entries denote self-comparisons.

The vehicle-related datasets (Figure~\ref{similar_data-plot}) form a tightly connected group, exhibiting high inter-coverage and structural overlap. In contrast, cross-domain comparisons (Figure~\ref{dissimilar_data-plot}) such as between healthcare, music, and automotive datasets show low feature alignment.

Together, these visualisations support our core observation: while some dataset groups exhibit redundancy and high overlap, the broader benchmark pool lacks diversity and coherent structure, underscoring the need for more principled dataset selection in text-tabular benchmarks.

\paragraph{Directional Table Comparisons.}
Tables~\ref{table:similar_counts} and ~\ref{table:dissimilar_counts} provide the raw counts underlying the directional similarity scores in Figures~\ref{similar_data-plot} and~\ref{dissimilar_data-plot}. Each cell reports the number of semantically matched features divided by the total number of features in the source dataset (row).

These counts demonstrate why directionality is essential. A large dataset may match many features in a smaller one, but this overlap may still represent a small fraction of its own schema. For example, \texttt{cars\_24} matches 2 features in \texttt{museums}, yielding 2/17 in one direction and 2/9 in the other—highlighting a twofold difference in perceived similarity.

We observe similar effects even among closely related datasets. In Table~\ref{table:similar_counts}, comparisons with \texttt{cardekho} (17 features) consistently yield lower scores from other datasets, despite sharing similar content. Without directionality, such asymmetries would be obscured, potentially misleading benchmark interpretations.

\begin{table}[!h]
\centering
\caption{Matched feature counts for dissimilar datasets. Row = source dataset, column = target dataset.}
\vskip 0.05in
\begin{tabular}{lccccccc}
\toprule
 & cars\_24 & museums & spotify & prescript\_drugs & wine\_vivino\_rating & videogame\_sales & anime\_planet \\
\midrule
cars\_24 & – & 2 / 17 & 3 / 19 & 3 / 8 & 3 / 9 & 2 / 6 & 2 / 11 \\
museums & 2 / 9 & – & 1 / 19 & 2 / 8 & 2 / 9 & 2 / 6 & 3 / 11 \\
spotify & 3 / 9 & 1 / 17 & – & 3 / 8 & 3 / 9 & 1 / 6 & 2 / 11 \\
prescript\_drugs & 3 / 9 & 2 / 17 & 3 / 19 & – & 4 / 9 & 2 / 6 & 3 / 11 \\
wine\_vivino\_rating & 3 / 9 & 2 / 17 & 3 / 19 & 4 / 8 & – & 2 / 6 & 4 / 11 \\
videogame\_sales & 2 / 9 & 2 / 17 & 1 / 19 & 2 / 8 & 2 / 9 & – & 4 / 11 \\
anime\_planet & 2 / 9 & 3 / 17 & 2 / 19 & 3 / 8 & 4 / 9 & 4 / 6 & – \\
\bottomrule
\end{tabular}

\label{table:dissimilar_counts}
\end{table}

\begin{table}
\centering
\caption{Matched feature counts for similar datasets. Row = source dataset, column = target dataset.}
\vskip 0.05in
\resizebox{\textwidth}{!}{
\begin{tabular}{lcccccccc}
\toprule
 & bikedekho & cardekho & cars\_24 & cars\_pakistan & bikewale & cars\_italy & cars\_SA & cars\_dot\_com \\
\midrule
bikedekho & – & 4 / 16 & 7 / 9 & 5 / 8 & 7 / 8 & 4 / 8 & 7 / 11 & 7 / 12 \\
cardekho & 4 / 8 & – & 7 / 8 & 4 / 8 & 5 / 8 & 5 / 8 & 8 / 11 & 7 / 11 \\
cars\_24 & 7 / 8 & 7 / 17 & – & 5 / 8 & 6 / 8 & 5 / 8 & 7 / 11 & 7 / 11 \\
cars\_pakistan & 5 / 8 & 4 / 17 & 5 / 9 & – & 5 / 8 & 4 / 8 & 6 / 11 & 7 / 11 \\
bikewale & 7 / 8 & 5 / 17 & 6 / 9 & 5 / 8 & – & 6 / 9 & 5 / 11 & 4 / 11 \\
cars\_italy & 4 / 8 & 5 / 17 & 5 / 8 & 4 / 8 & 6 / 8 & – & 5 / 11 & 4 / 11 \\
cars\_SA & 7 / 8 & 8 / 17 & 7 / 9 & 6 / 8 & 5 / 8 & 5 / 8 & – & 9 / 11 \\
cars\_dot\_com & 7 / 8 & 7 / 17 & 7 / 9 & 7 / 9 & 4 / 8 & 4 / 8 & 9 / 12 & – \\
\bottomrule
\end{tabular}
}

\label{table:similar_counts}
\end{table}

\subsection{Dataset-Specific Pre-processing}\label{sec:dataset-preproc}

A less visible, but highly consequential, aspect of \textsc{CARTE} is the amount of \emph{dataset-specific} pre-processing embedded in the official dataset loaders shared by the authors. 
These routines do far more than generic cleaning: while many datasets receive only light scrubbing, a non-trivial subset is significantly modified: columns are renamed, features merged or split, categorical values recoded with expert priors, and even the target variable transformed.
While such changes may boost model performance by providing richer semantics, the paper itself mentions only “minimal preprocessing,” without elaboration. 
Because these steps are hard-coded and undocumented, reproducing \textsc{CARTE} results with an independent pipeline is effectively impossible.

\textbf{Not all datasets are affected.}  
While many datasets in \textsc{CARTE} undergo only minimal hygiene steps (e.g., dropping high-null columns or coercing obvious types), we find that a substantial subset, roughly 10 out of the 51, are subject to heavier, dataset-specific transformations.  
These go beyond what we would consider acceptable for a general-purpose benchmark and instead introduce domain-specific engineering choices.  
Our concern is not with basic cleanup, but with the lack of transparency and consistency in how deeper interventions are applied.  
For details on what we consider minimal and acceptable preprocessing, see Section~\ref{appendix-datasets}.

\paragraph{Illustrative examples.}
Listings~\ref{lst:whisky_preproc} and~\ref{lst:pakistan_preproc} show two typical cases:

\begin{itemize}[leftmargin=1.8em,itemsep=0.25em]
\item \textbf{\texttt{whisky}}  
  \begin{enumerate*}
    \item maps a five-level price code (\$–\$\$\$\$\$) to coarse text ranges,
    \item replaces a one-letter flavour \emph{cluster} with a multi-word description,
    \item drops three columns, fills missing values, binarises the {\small\texttt{Meta\_Critic}} target at 8.6, and
    \item prunes high-null and single-unique features.
  \end{enumerate*}
\item \textbf{\texttt{used\_cars\_pakistan}}  
  \begin{enumerate*}
    \item concatenates {\small\texttt{Brand}}, {\small\texttt{Model}}, and {\small\texttt{Version}} into one string,
    \item log-transforms the price (base 100),
    \item coerces several numeric columns to strings or floats, and
    \item applies the same null- and uniqueness-based column drops.
  \end{enumerate*}
\end{itemize}

\begin{lstlisting}[language=Python,
  caption={CARTE's key preprocessing steps for \texttt{whisky}},
  label={lst:whisky_preproc}]
# --- recode price brackets (5 -> 6 text ranges) ------------------
data["Cost"] = data["Cost"].map({
    "$$$$$+": "over 300 CAD",
    "$$$$$" : "125-300 CAD",
    "$$$$"  : "70-125 CAD",
    "$$$"   : "50-70 CAD",
    "$$"    : "30-50 CAD",
    "$"     : "< 30 CAD"
})

# --- expand one-letter flavour clusters -------------------------
cluster_map = {"A": "... sweet, fruity, spicy",
               "B": "... floral, malty",
               "J": "dry, very smoky, pungent", ...}
data["Cluster"] = data["Cluster"].map(cluster_map)

# --- binarise target -------------------------------------------
data["Meta_Critic"] = (data["Meta_Critic"] > 8.6).astype(int)
\end{lstlisting}

\begin{lstlisting}[language=Python,
  caption={CARTE's key preprocessing steps for \texttt{used\_cars\_pakistan}},
  label={lst:pakistan_preproc}]
# --- standardise column names & dtypes -------------------------
data.rename(columns={"Make":"Brand","Make_Year":"Year","CC":"Engine_Capacity"}, inplace=True)
data = data.astype({"Year":str, "Engine_Capacity":float, "Mileage":float})

# --- join brand/model/version into one text feature ------------
data["Model"] = data["Brand"] + " " + data["Model"] + ", " + data["Version"]
data.drop(columns=["Brand","Version"], inplace=True)

# --- log-transform target (base-100) ---------------------------
data["Price"] = np.emath.logn(100, data["Price"])
\end{lstlisting}

\paragraph{Why this matters.}
Such tailored transformations effectively inject domain knowledge and can inflate performance for models that rely on these engineered signals (e.g., composite model names or price buckets).  
At the same time, other models that might have performed better with the original data are negatively represented. 
We argue that one would either perform manual preprocessing for all models individually, or for none at all. 

\subsection{Artificial Targets and Label Construction}
\label{sec:artificial-targets}

Another concern in \textsc{CARTE} is the presence of \emph{artificial or weakly motivated targets}, particularly in their classification tasks. Of the 11 classification datasets, many are derived from regression settings where continuous labels are simply thresholded into binary classes.

This design choice introduces several problems:
\begin{itemize}
    \item \textbf{Loss of task fidelity:} Binarising a continuous outcome (e.g., review scores or numerical metrics) removes useful information and distorts the underlying distribution.

    \item \textbf{Unclear target motivation:} In several cases, the classification target does not appear meaningfully aligned with the available features. For example, in \texttt{nba\_draft}, the label indicates whether a player had “positive value over replacement”. This is a long-term outcome influenced by team context, opportunity, and external factors not represented in the dataset. Without such contextual features, it’s unclear how this target can be reliably predicted.

    \item \textbf{Reduced benchmark utility:} These constructions skew the task distribution toward binary classification, limiting the benchmark's diversity and making it a poor proxy for real-world classification challenges.
\end{itemize}

Not all classification targets in \textsc{CARTE} are problematic, e.g., \texttt{spotify} uses “major vs minor key” which, while narrow, is at least categorical by nature. But overall, we find that most of the classification datasets do not represent native classification problems and instead reflect arbitrary thresholding over numerical data.

For a robust benchmark, classification tasks should arise from semantically grounded, categorical outcomes rather than being retrofitted from regression labels.

\subsection{Data Accessibility and Reproducibility}
\label{sec:data-access}

While not a primary concern for \textsc{CARTE}, we believe it's important to highlight a broader issue in multimodal tabular benchmarks: \emph{dataset accessibility}. For a benchmark to be truly useful and reproducible, its datasets should be easy to access, ideally via direct download or a minimal sign-up process.

Unfortunately, several works in this domain, such as the AutoML Benchmark \citep{autogluon-text-tabular-benchmark}, include datasets sourced from platforms like \href{https://machinehack.com}{MachineHack} or private competitions. These often require non-trivial setup: creating platform accounts, joining specific challenges, or navigating manual approval processes.

The situation is even more restrictive for medical datasets like \texttt{MIMIC-III}, where individual institutional review and data use agreements are needed. While these restrictions are understandable in domain-specific contexts, they pose a significant barrier for community benchmarks that aim to support widespread experimentation and reproducibility.

We emphasize that we have no issues with commonly used sources such as Kaggle, OpenML, UCI, or direct download via \texttt{wget}. These platforms offer broad accessibility and minimal setup, making them reasonable choices for benchmark construction. The real challenge arises when datasets require significantly more effort to access—such as navigating opaque approval processes, participating in gated competitions, or relying on unstable or temporary hosting. Such sources are inherently less reliable; datasets may become inaccessible without notice, undermining the longevity and reproducibility of benchmark results.

In our view, benchmark datasets should be:
\begin{itemize}
\item publicly accessible without approval steps;
\item accompanied by scripts or loaders that work out of the box
\item stable over time (i.e., not dependent on ephemeral hosting or event-based portals).
\end{itemize}

Without these guarantees, reproducing results becomes cumbersome, and sharing new methods depends on access to fragile or gated data sources, undermining the core purpose of open and lasting benchmarks.

\subsection{Dual Signal}
\label{sec:dual-signal}

An important design consideration in our benchmark was ensuring that each dataset provides a meaningful \emph{dual signal}—that is, both the structured and textual features contribute useful, non-redundant information to the prediction task. We aimed to avoid cases where the target could be effectively predicted from only one modality, rendering the other superfluous.

This consideration became particularly relevant when reviewing datasets from the AutoML benchmark~\citep{autogluon-text-tabular-benchmark}, which includes examples such as \texttt{qaa} and \texttt{qaq}, derived from Google's question-answer corpus. These datasets contain textual features like \texttt{question}, \texttt{question title}, and \texttt{answer}, alongside a single categorical feature (\texttt{question category}). While valid for multimodal learning, their structure arguably makes them better suited for NLP-style modeling than for evaluating methods rooted in tabular learning.

Although we do not quantify dual signal in this paper, it served as a key guiding principle in our dataset selection. A formal analysis of how predictive signal is distributed across structured features and textual embeddings would provide stronger support for this criterion, but was beyond the scope of this work and is left to future studies.

\subsection{Free Text Is Not Always Free Text}
\label{sec:text-not-categorizable}

When building benchmarks for tabular models that incorporate textual inputs, it is essential to verify that the so-called "text" columns are truly \emph{semantic} in nature, not merely strings in format. Many datasets include string-valued fields that function more like categorical variables or identifiers, such as product codes, model names, or brand labels. While technically textual, these features typically do not require or benefit from text-aware modeling. 

To better characterize this, our internal analysis includes basic statistics such as the number of unique values per column. This helps reveal whether string columns behave more like free text or structured categories. For instance, a column with only a few dozen repeated string values is unlikely to encode the kind of nuanced semantics that justify using advanced text encoders.

In the case of \textsc{CARTE}, to be fair, the benchmark does not explicitly position itself as targeting free-text reasoning. However, it has often been interpreted that way within the community. Based on our analysis, only around half of the datasets (depending on the threshold applied) contain string features with enough diversity to be meaningfully treated as textual inputs. This highlights the importance of validating the modality of features, rather than assuming all string-typed columns warrant semantic modeling.

Yet, at least one example of a dataset with non-semantically rich free text features: `us-accidents-counts' dataset in the CARTE benchmark has textual columns such as 'ZIPCODE' or `AIRPORT-CODE' which have no semantic value or concrete relevance to the target feature of accident counts in the city. 

\subsection{How to Choose a Good Prediction Task Dataset?} 
To assess the suitability of datasets for prediction tasks, we conducted an analysis on both CARTE's benchmark datasets and those in our own benchmark. This analysis leveraged GPT-4o to evaluate whether each dataset is well-aligned with the intended task objectives. The prompt used for this evaluation is shown in Listing~\ref{lst:data-fit-prompt}.

We illustrate GPT's assessment using two example datasets: CARTE's \texttt{us-accidents} and our own \texttt{beer-rating}. Based on GPT's evaluations, the majority of CARTE's datasets were categorized as either \textcolor{red}{Red} or \textcolor{yellow}{Yellow}, indicating potential issues in task alignment. In contrast, our benchmark datasets were predominantly classified as \textcolor{green}{Green}, suggesting a stronger fit for the intended prediction tasks.

These findings support our motivation to develop a new benchmark that addresses some of CARTE's limitations, as identified through GPT's interpretations. It is important to note that these assessments were based on GPT-4o at the time of testing and may evolve with future model updates. 

\begin{lstlisting}[language=Python, numbers=none, caption={Prompt for analysing dataset fit for Text for Tabular data benchmark.}, label={lst:data-fit-prompt}]
prompt_template = """
I am a researcher in the field of AI and I want to create a benchmark for tabular datasets with meaningful textual features. The textual features would be replaced with textual embeddings and the dataset will be used to benchmark a set of different tabular models.  For each dataset I will provide you with column names, first few rows and declaration of the target feature. Based on conditions below, review each dataset and classify it as Green (meets all conditions), Yellow (meets some), or Red (meets none) based on its fitness to be included in the benchmark, then justify your choice. 
The general conditions are: 
(1) suitability for regression/classification (not recommendation or look up table tasks), 
(2) prediction is to be boosted by both textual and non textual features 
(3) target feature is native to the prediction task and relavant to the feature space
(4) Textual features are semantically rich (e.g., 'item_condition' > 'seller_name').
(5) The features must contain enough signal for the model to go beyond predicting the mean/target statistics.
Always explain your reasoning 
(e.g., 'Yellow: meets 1, 3, and 5 but lacks long-text features' and hence is not a good fit for the benchmark).
"""
\end{lstlisting}

\begin{lstlisting}[language=Python, numbers=none, caption={GPT's verdict on `us-accidents-counts' dataset from CARTE benchmark.}]
Answer: """
Classification: RED

Analysis:

Condition 1: Suitability for regression or classification

Met. The target column "Counts" is a numeric value, which makes the dataset suitable for a regression task.

Condition 2: Prediction is to be boosted by both textual and non-textual features

Not met. While the dataset includes textual features like "City", "County", and "State", these are short categorical labels. They are not semantically rich and are unlikely to benefit from textual embeddings. Features like "Zipcode", "Code", and "Airport_Code" are identifiers or codes, not meaningful text.

Condition 3: Target is native to the task and relevant to the features

Partially met. The target "Counts" might represent some countable event or measurement, but there is no clear link between the features and the target. For example, the same city and zipcode appear with very different count values, suggesting missing context.

Condition 4: Textual features are semantically rich

Not met. The textual features in the dataset are names of locations and codes. These are not descriptive, long-form, or semantically meaningful in a way that text embeddings would capture. This limits their utility in a benchmark focused on rich textual input.

Condition 5: Dataset has enough signal to go beyond predicting the mean

Partially met. While 'Counts' varies across rows, the available features do not clearly explain that variance. This limits a model's ability to learn meaningful patterns beyond simple statistics.

Conclusion:
This dataset is not a good candidate for a benchmark focused on textual embeddings in tabular models. It lacks rich textual features and does not offer strong predictive links between inputs and target values.

Final Rating: RED: Only condition 1 is clearly met.
"""
\end{lstlisting}

\begin{lstlisting}[language=Python, numbers=none, caption={GPT's verdict on `beer-rating' dataset from our benchmark.}]
Answer: """
Classification: GREEN

Analysis:

Condition 1: Suitability for regression or classification
Met. The target column "review_overall" is a numeric value, which makes this a valid regression task.

Condition 2: Prediction is to be boosted by both textual and non-textual features
Met. The dataset includes a rich textual feature ("Description") that contains tasting notes and product details. In addition, there are structured numeric features like "ABV", "Min IBU", "Max IBU", and flavor attributes such as "Sweet", "Sour", and "Hoppy". This combination supports modeling with both types of features.

Condition 3: Target is native to the task and relevant to the feature space
Met. The "review_overall" rating is directly related to the beer's style, taste, and description. It is a natural outcome of the given features, which makes it highly relevant.

Condition 4: Textual features are semantically rich
Met. The "Description" field includes full sentences and tasting notes. This is semantically rich text that can benefit from text embeddings and contributes meaningfully to model performance.

Condition 5: Dataset has enough signal to go beyond predicting the mean
Met. There is variability in the review scores and the number of reviews. Combined with detailed feature data, the model has enough signal to learn meaningful patterns rather than just guessing the average.

Conclusion:
This dataset is well-suited for benchmarking tabular models that integrate textual embeddings. It includes a strong combination of structured and unstructured features, a meaningful target, and enough variability to support learning.

Final Rating: GREEN - Meets all conditions.
"""
\end{lstlisting}

\subsection{Final Statement for Prior Benchmarks}
\label{sec:benchmark-appreciation}

While we have pointed out several limitations of the prior benchmarks, we also acknowledge their value. These works played a crucial role in shaping our design decisions and served as a practical foundation for our dataset selection.

Several datasets from these earlier benchmarks were included in our final benchmark, in some cases after minor modifications (such as using a different Kaggle version) to ensure consistency and usability. In addition to our main selection, we also provide a number of other datasets in our \href{https://github.com/mrazmartin/TextTabBench/tree/main}{GitHub repository} that did not fully meet our inclusion criteria but are still worth sharing. These near-threshold datasets may fall short of our stricter standards for signal quality, diversity, yet they can still be useful for broader experimentation.

\end{document}